\documentclass[10pt,twocolumn,letterpaper]{article}
\usepackage[pagenumbers]{wacv}

\usepackage{float}
\usepackage{enumitem}

\usepackage{amsmath}
\usepackage{amssymb}
\usepackage{amsfonts}
\usepackage{bm}

\usepackage{algorithm}
\usepackage{algpseudocode}
\algrenewcomment[1]{\hfill$\triangleright$~#1}
\algrenewcommand\algorithmicrequire{\textbf{Input:}}
\algrenewcommand\algorithmicensure{\textbf{Return:}}

\usepackage{multirow}
\usepackage{booktabs}
\usepackage{graphicx}
\usepackage{subcaption}
\usepackage{arydshln}

\usepackage{wrapfig}

\usepackage[table, dvipsnames]{xcolor}

\usepackage{thmtools}
\usepackage{thm-restate}

\usepackage{titletoc}

\usepackage[most]{tcolorbox}

\definecolor{DarkGreen}{rgb}{0.00, 0.40, 0.00}
\definecolor{RoyalBlue}{rgb}{0.25, 0.35, 0.74}
\definecolor{Crimson}{rgb}{0.86, 0.08, 0.24}
\definecolor{ChromeYellow}{rgb}{1.0, 0.65, 0.0}

\usepackage{array}
\usepackage{xcolor}

\newcolumntype{+}{>{\global\let\currentrowstyle\relax}}
\newcolumntype{^}{>{\currentrowstyle}}

\definecolor{noncomparablegray}{RGB}{145,145,145}

\usepackage[table, dvipsnames]{xcolor}

\tcolorboxenvironment{definition}{
  colback=blue!3,
  colframe=blue!3,
  boxrule=0.5pt,
  arc=6pt,
  left=6pt,
  right=6pt,
  top=5pt,
  bottom=5pt,
  before skip=6pt,
  after skip=6pt
}

\tcolorboxenvironment{lemma}{
  colback=blue!3,
  colframe=blue!3,
  boxrule=0.5pt,
  arc=6pt,
  left=6pt,
  right=6pt,
  top=5pt,
  bottom=5pt,
  before skip=6pt,
  after skip=6pt
}

\tcolorboxenvironment{theorem}{
  colback=blue!3,
  colframe=blue!3,
  boxrule=0.5pt,
  arc=6pt,
  left=6pt,
  right=6pt,
  top=5pt,
  bottom=5pt,
  before skip=6pt,
  after skip=6pt
}

\usepackage{tocloft}

\newcommand\DoToC{%
  \startcontents
  {\bfseries \noindent Table of Contents\par}
  \vskip5pt
  \printcontents{}{1}{\hrule\vskip5pt}
  \vskip5pt\hrule\vskip5pt
}

\usepackage{amsmath}
\usepackage{amssymb}
\usepackage{amsthm}
\usepackage{amsfonts}
\usepackage{bm}
\usepackage{pifont}
\newcommand{\xmark}{\ding{55}}

\usepackage{multirow}
\usepackage{booktabs}
\usepackage{graphicx}
\usepackage{subcaption}
\usepackage{arydshln}
\usepackage{makecell}

\newtheorem{theorem}{Theorem}
\newtheorem{lemma}{Lemma}

\definecolor{DarkGreen}{rgb}{0.00, 0.40, 0.00}
\definecolor{RoyalBlue}{rgb}{0.25, 0.35, 0.74}
\definecolor{Crimson}{rgb}{0.86, 0.08, 0.24}
\definecolor{ChromeYellow}{rgb}{1.0, 0.65, 0.0}

\newcommand{\method}{QVR3D}
\colorlet{ourshl}{blue!10}
\newcommand{\first}[1]{\textbf{#1}}
\newcommand{\second}[1]{\underline{#1}}
\newcommand{\argmax}{\operatornamewithlimits{argmax}}

\usepackage[most]{tcolorbox}
\tcbuselibrary{listings,breakable}
\newtcblisting{promptbox}[1][]{
    listing only,
    title={#1},
    fonttitle=\bfseries,
    listing options={
        basicstyle=\ttfamily\small,
        breaklines=true,
        breakindent=0pt,
        breakautoindent=false,
        columns=fullflexible
    }
}

\definecolor{wacvblue}{rgb}{0.21,0.49,0.74}
\usepackage[pagebackref,breaklinks,colorlinks,allcolors=wacvblue]{hyperref}

\title{QVR3D: Query-Guided Visual Reasoning for Training-Free 3D Grounding}

\author{
Jung-Hao Kao\textsuperscript{1,3}\thanks{Equal contribution.}
\qquad
Jing-En Huang\textsuperscript{2,3}$^{*}$
\qquad
Jia-Wei Liao\textsuperscript{1,3}$^{*}$\thanks{Project leader.}
\qquad
Tzuhsuan Huang\textsuperscript{3} \\
Yu-Lun Liu\textsuperscript{2}
\qquad
Cheng-Fu Chou\textsuperscript{1}
\qquad
Jun-Cheng Chen\textsuperscript{3}\thanks{Corresponding author.}
\vspace{0.5em} \\
\textsuperscript{1}National Taiwan University \quad
\textsuperscript{2}National Yang Ming Chiao Tung University \\
\textsuperscript{3}Research Center for Information Technology Innovation, Academia Sinica
}

\begin{document}
\makeatletter
\let\wacvOriginalMaketitle\@maketitle
\renewcommand{\@maketitle}{%
    \wacvOriginalMaketitle
    \begin{center}
        \includegraphics[width=\linewidth]{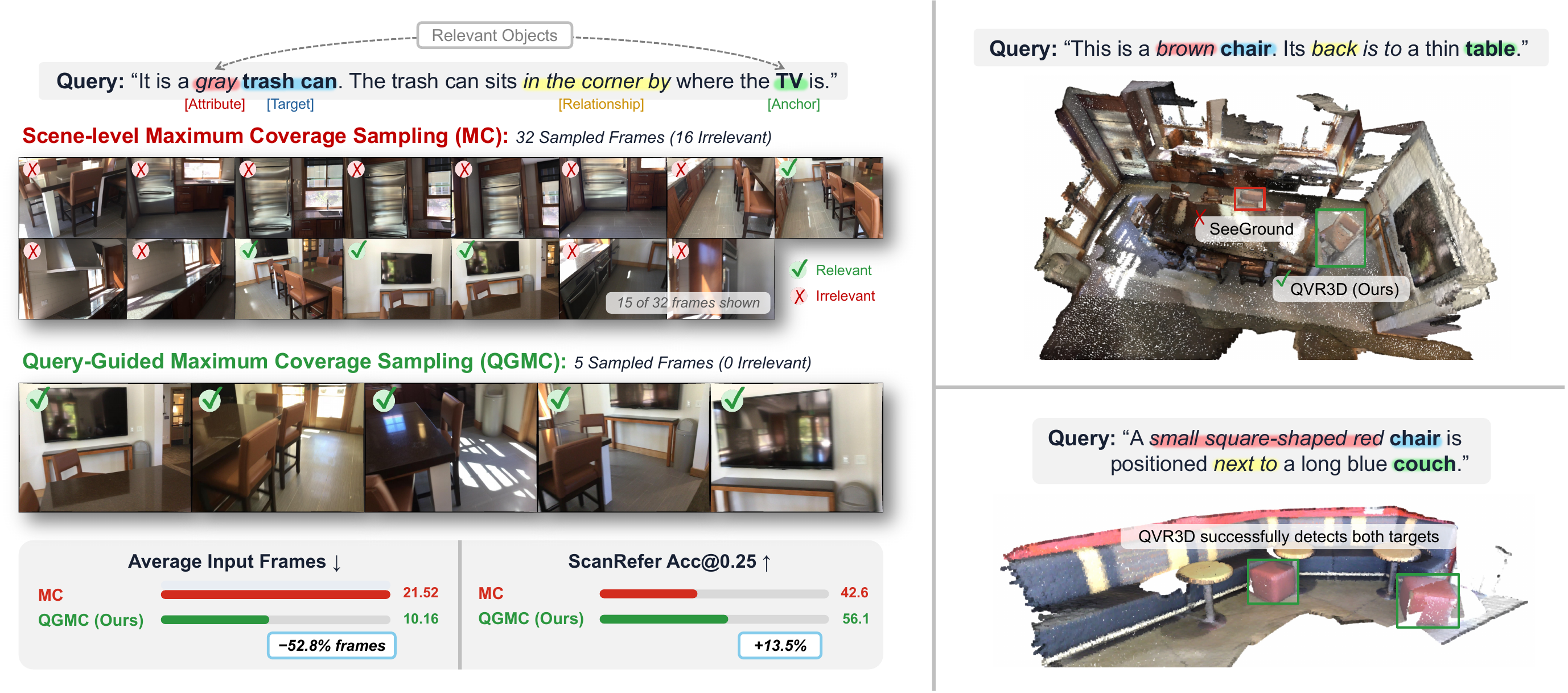}
        \captionsetup{hypcap=false}
        \captionof{figure}{\textbf{Overview of QVR3D's query-guided frame selection and 3D grounding capabilities.}
        \textit{(Left)} Compared with the scene-level maximum coverage sampling (MC), our proposed query-guided maximum coverage sampling (QGMC) instead leverages the target and anchor categories parsed from the query to construct query-relevant regions for frame sampling. Our method not only retrieves a more compact number of relevant frames but also achieves better performance on the ScanRefer dataset. \textit{(Upper right)} The queried spatial relation (``the chair has its back to a thin table'') distinguishes the target from other chairs of the same category. SeeGround selects a distractor (red box), whereas QVR3D correctly localizes the chair consistent with this relation (green box). \textit{(Lower right)} On Multi3DRefer, QVR3D localizes both chairs satisfying the description (green boxes), demonstrating multi-target grounding.}
        \label{fig:teaser}
    \end{center}
}
\makeatother
\maketitle

\begin{abstract}
3D visual grounding localizes objects in 3D scenes from natural language descriptions, requiring reasoning over semantics, spatial relationships, and multi-view observations. Training-free approaches avoid costly annotations but often rely on explicit scene transformation or scene-level frame selection, which may waste limited visual capacity on query-irrelevant regions. We observe that effective grounding should prioritize observations that provide evidence for the query rather than uniformly represent the entire scene. Based on this insight, we introduce QVR3D, a query-guided visual reasoning framework that operates on RGB-D observations using off-the-shelf foundation models. QVR3D combines Query-Guided Maximum Coverage (QGMC) for selecting complementary query-relevant views with Grounding Probability Estimation (GPE) that integrates instance confidence, attribute consistency, and relationship consistency. The framework supports both single- and multi-object grounding without task-specific adaptation. Experiments on ScanRefer and Nr3D achieve state-of-the-art zero-shot performance, while QGMC also improves frame efficiency across multiple 3D vision-language tasks. Our code is available at \url{https://github.com/jwliao1209/QVR3D}.
\end{abstract}

\section{Introduction}

3D visual grounding~\cite{chen2020scanrefer,achlioptas2020referit3d} aims to localize objects in a 3D scene from natural language descriptions, providing a fundamental interface between language understanding and spatial perception. This capability is essential for embodied applications such as robotics, autonomous navigation, human-computer interaction, and augmented reality, where an agent must identify objects not only by appearance but also by their spatial context and relationships to surrounding objects. Despite the strong visual understanding capabilities of recent vision-language models (VLMs)~\cite{wang2024qwen2,bai2025qwen3,chen2024internvl}, extending grounding from 2D images to 3D environments remains challenging because the model must reason over geometry, object relationships, and observations distributed across multiple viewpoints.

Existing 3D visual grounding methods primarily follow supervised~\cite{chen2020scanrefer,yuan2021instancerefer,zhao20213dvg,huang2022multi,luo20223d,wu2023eda,wang2024g,qian2024multi,unal2024four,jia2024sceneverse,guo2025tsp3d,tan2026pc,wang2023distilling} or training-free~\cite{yang2024llm,yuan2024visual,li2025seeground,liu2026view,drozdov2026z3d} paradigms. Supervised approaches learn language-to-geometry alignment from large-scale annotated 3D data, but acquiring object-level grounding annotations is expensive and difficult to scale. Recent training-free methods instead leverage pretrained foundation models without task-specific optimization: they typically convert 3D scenes into rendered views, textual spatial representations, or structured scene graphs that can be processed by 2D VLMs. While effective, these approaches often require explicit scene transformation and are largely designed for single-object grounding. Moreover, frame-based 3D vision-language systems typically select observations according to global scene coverage. Under a limited frame budget, maximizing scene-level coverage can allocate substantial capacity to regions unrelated to the language query, leaving fewer informative observations for resolving the referred object. This raises a key question: \textit{can training-free 3D grounding reason directly from RGB-D observations while allocating visual evidence according to the query rather than the entire scene?}

Motivated by this need, we introduce \emph{QVR3D}, a query-guided visual reasoning framework for training-free 3D grounding, as illustrated in Fig.~\ref{fig:teaser}. Our key observation is that effective grounding does not require uniformly representing the entire scene; instead, the visual budget should concentrate on objects and regions that provide evidence for the query. QVR3D therefore operates directly on RGB-D video observations and detected 3D instances without additional scene rendering or task-specific adaptation. It first identifies candidate target and anchor objects from the language query and introduces \emph{Query-Guided Maximum Coverage (QGMC)} to select complementary views that cover query-relevant regions. The selected observations are then used by \emph{Grounding Probability Estimation (GPE)} to combine instance confidence with attribute and relationship consistency, providing complementary evidence for VLM-based grounding selection. This formulation naturally handles both single- and multi-object queries within the same inference pipeline.

Extensive experiments on ScanRefer~\cite{chen2020scanrefer} and Nr3D~\cite{achlioptas2020referit3d} demonstrate strong training-free 3D grounding performance. QVR3D achieves state-of-the-art results among zero-shot methods on ScanRefer and remains competitive with several supervised approaches. Across challenging splits involving multiple same-category objects, our results show that concentrating visual observations on query-relevant regions improves fine-grained object disambiguation. We evaluate QGMC as a plug-and-play sampling strategy for frame-based 3D vision-language systems and observe consistent improvements across visual grounding, dense captioning, and question answering, indicating that query-guided observation selection generalizes beyond the grounding task. Moreover, as shown in Fig.~\ref{fig:teaser}, scene-level Maximum Coverage (MC) selects 32 frames, including 16 irrelevant views, whereas the QGMC of our QVR3D selects 5, all query-relevant. On ScanRefer with Qwen3-VL-8B in the frame-only setting without GPE, QGMC reduces the average frames from 21.52 to 10.16 (-52.8\%), and improves Acc@0.25 from 42.6\% to 56.1\% (+13.5\%). Overall, our QVR3D achieves 56.9\% with all the proposed modules, demonstrating the effectiveness of the proposed approach.

Our contributions are summarized as follows:
\begin{enumerate}
\item We introduce QVR3D, a query-guided visual reasoning framework for training-free 3D grounding that directly operates on RGB-D observations using off-the-shelf VLMs without task-specific adaptation.

\item We propose Query-Guided Maximum Coverage (QGMC), which prioritizes query-relevant regions for efficient frame selection and admits a monotone submodular objective with theoretical approximation guarantee. QGMC is plug-and-play and improves multiple frame-based 3D vision-language tasks.

\item We develop Grounding Probability Estimation (GPE), which integrates instance confidence, attribute consistency, and relationship consistency to provide complementary grounding cues for VLM-based target selection.

\item Our framework supports both single- and multi-object grounding. Extensive experiments on ScanRefer and Nr3D demonstrate state-of-the-art zero-shot performance and strong generalization across settings.
\end{enumerate}

\section{Related Work}

\subsection{3D Vision-Language Scene Understanding}
Recent vision-language models (VLMs)~\cite{wang2024qwen2,bai2025qwen3,chen2024internvl} have been extended from 2D images to 3D environments by incorporating geometric structure, multi-view observations, or video representations. These models support a range of 3D vision-language tasks, including question answering~\cite{azuma2022scanqa}, dense captioning~\cite{chen2021scan2cap}, and visual grounding~\cite{chen2020scanrefer,achlioptas2020referit3d}. Video-3D LLM~\cite{zheng2025video}, for example, represents a 3D scene as a video sequence and incorporates 3D positional encoding to preserve spatial information across views. It also introduces maximum coverage sampling to reduce redundant observations while maintaining scene coverage. Despite their strong performance, such 3D VLMs typically rely on large-scale instruction tuning~\cite{liu2023visual} and substantial computational resources. Our work instead focuses on training-free 3D grounding with off-the-shelf foundation models and directly operates on RGB-D observations without adaptation.

\subsection{3D Visual Grounding}

\paragraph{Supervised Methods.}
Supervised 3D visual grounding methods~\cite{yuan2021instancerefer, huang2021text,yang2021sat,chen2022language,luo20223d,jain2022bottom,chang2024mikasa,wang2024g,qian2024multi,wang2025liba,guo2025tsp3d,tan2026pc,tan2026uz3dvg} learn language-to-geometry alignment from annotated 3D scenes. TSP3D~\cite{guo2025tsp3d} improves grounding efficiency through text-guided pruning, which suppresses irrelevant voxel features, together with completion-based addition to recover useful geometric information. PC-CrossDiff~\cite{tan2026pc} models cross-modal interactions at both point and cluster levels using differential attention to enhance localization cues and suppress spatial ambiguity. While supervised methods achieve strong performance, they require task-specific 3D annotations and training, motivating approaches that can directly leverage pretrained foundation models.

\begin{figure*}
    \centering
    \includegraphics[width=\linewidth]{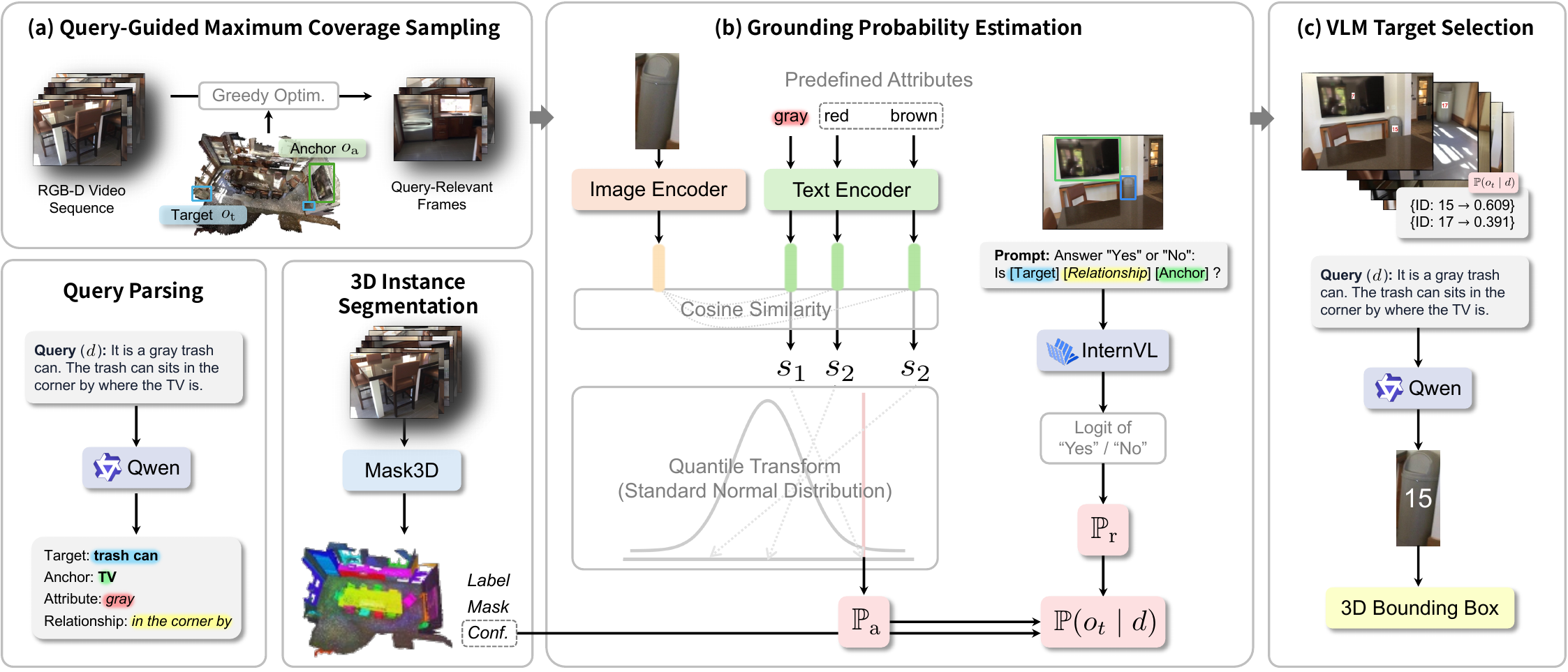}
    \caption{\textbf{Overview of the QVR3D pipeline.}
    Given an RGB-D sequence and a natural language query $d$, QVR3D parses target and anchor cues and extracts 3D object instances. (a) Query-Guided Maximum Coverage Sampling (QGMC) selects complementary query-relevant frames by maximizing the coverage of candidate target and anchor voxels. (b) Grounding Probability Estimation (GPE) combines instance confidence, attribute consistency, and relationship consistency to estimate $P(o_{t}\mid d)$. (c) VLM Target Selection uses the selected frames, query, candidate IDs, and grounding probabilities to identify target instances and retrieve their 3D bounding boxes.
    }
    \label{fig:pipeline}
\end{figure*}

\paragraph{Training-Free Methods.}
Training-free 3D grounding methods~\cite{li2025seeground, mi2025language,cai2025naver,liu2026view,drozdov2026z3d} avoid task-specific training by transforming or organizing 3D scene information into representations that pretrained language and vision-language models can process. SeeGround~\cite{li2025seeground} converts a 3D scene into query-aligned multi-view renderings and combines them with explicit spatial descriptions for VLM-based grounding. Language-to-Space Programming (LaSP)~\cite{mi2025language} formulates grounding as language-guided spatial reasoning and uses LLM-generated programs to analyze relationships among detected objects. View-on-Graph (VoG)~\cite{liu2026view} represents the scene as a multimodal graph and enables the model to retrieve and reason over relevant nodes and relations. These approaches demonstrate that pretrained foundation models can perform 3D grounding without task-specific training, but generally rely on rendered views, executable programs, or structured scene representations. Our framework instead operates directly on RGB-D observations and detected 3D instances, avoiding additional scene rendering or explicit scene conversion. Existing training-free approaches also primarily target single-object grounding, whereas our framework supports queries referring to multiple target instances within the same inference pipeline.

\subsection{Visual Observation Selection for 3D Reasoning}
Frame-based 3D vision-language systems must select limited observations from long RGB-D sequences. Uniform frame sampling is computationally simple but ignores scene content and may retain redundant observations. Coverage-based strategies address this issue by selecting views that collectively observe a larger portion of the scene. Video-3D LLM~\cite{zheng2025video}, for example, applies maximum coverage sampling over scene voxels to improve frame utilization. IVSGround~\cite{chiang2026look} learns to select query-relevant views using VLM-generated feedback for 3D grounding.

Scene-level coverage is independent of the language query, so selected frames may devote limited capacity to regions unrelated to the referred objects. This motivates query-aware observation selection that prioritizes visual evidence relevant to grounding. Our method follows this direction by selecting RGB-D observations according to query-relevant scene content, as described in Sec.~\ref{sec:qgmc}.

\section{Method}

\subsection{Overview}
Given an RGB-D video sequence $\mathcal{F} = \{f_1, f_2, ..., f_n\}$ captured in a static scene and a textual description referring to one or more target objects, our goal is to localize the described object(s) in 3D space. We first apply an off-the-shelf 3D instance segmentation model~\cite{schult2023mask3d} to obtain a set of object instances $\mathcal{O} = \{o_1, o_2, ..., o_N\}$. Each instance $o_i$ is associated with a voxel-level binary mask $m_i$, a semantic category label $y_i$, and a confidence score $c_i$. The task is then to identify the instance(s) that best match the textual description $d$ and return their corresponding 3D bounding boxes.

Our framework proceeds in four stages (Fig. \ref{fig:pipeline}). First, a large language model (LLM) parses the textual description into a target category, anchor categories, attributes, and spatial relationships, which are used to retrieve candidate target and anchor instances from $\mathcal{O}$. Second, we introduce Query-Guided Maximum Coverage (QGMC) sampling (Sec.~\ref{sec:qgmc}) to select informative RGB-D frames by maximizing the coverage of query-relevant objects rather than the entire scene. Third, our Grounding Probability Estimation (GPE) module uses the selected frames to estimate attribute and relationship consistency for each candidate target and combines these cues with the instance segmentation confidence to obtain its grounding probability (Sec.~\ref{sec:gpe}). Finally, the sampled frames, textual description, and estimated grounding probabilities are jointly provided to a vision-language model (VLM)~\cite{wang2024qwen2,bai2025qwen3,chen2024internvl}, which selects the object instance(s) that best match the query (Sec.~\ref{sec:target_selection}). The corresponding 3D bounding boxes are returned as the final grounding results.

\subsection{Query-Guided Maximum Coverage Sampling}
\label{sec:qgmc}

Recent training-free 3D grounding methods often convert a scene into bird's-eye-view (BEV) renderings or other compact visual representations before reasoning with a VLM. In contrast, we directly represent the scene using RGB-D frames sampled from the input video. Because modern VLMs can process only a limited number of frames, the choice of visual observations is critical. Existing frame sampling strategies typically maximize coverage of the entire scene, which preserves global context but may allocate much of the frame budget to regions unrelated to the language query. This introduces redundant visual content and reduces the amount of query-relevant evidence available to the VLM. To address this issue, we propose \textit{Query-Guided Maximum Coverage (QGMC) Sampling}, which concentrates the frame budget on candidate target and anchor objects derived from the query. Rather than maximizing coverage over all scene voxels, QGMC maximizes coverage only over query-relevant voxels, yielding a compact set of frames with more concentrated query-relevant evidence and less irrelevant context.

\paragraph{Query Parsing.}
Given a textual description $d$, we extract the semantic information required to identify the referred object(s). An LLM parses $d$ into four components: a target object category $t$, a set of anchor categories $\mathcal{A}= \{a_1, a_2, \ldots \}$ that provide contextual references, a set of attributes $\mathcal{P} =\{p_1, p_2, \ldots \}$ that describe properties of the target, and a set of relationships $\mathcal{R} = \{r_1, r_2, \ldots \}$ that characterize its relations to surrounding objects. For example, given the query ``It is a gray trash can. The trash can sits in the corner by where the TV is.'', the LLM identifies ``trash can'' as the target category, ``TV'' as an anchor category, ``gray'' as an attribute, and ``in the corner by TV'' as a relationship.

\paragraph{Candidate Object Retrieval.}
We next retrieve candidate target and anchor instances by matching the parsed object names against the semantic category labels predicted by the instance segmentation model. Because the parsed names may not exactly match the predicted labels due to wording differences, pluralization, or minor spelling variations, we combine fuzzy matching~\cite{lcvenshtcin1966binary} and stem matching~\cite{porter1980algorithm}. Fuzzy matching handles lexical and typographical variations through string similarity, while stem matching captures grammatical variants that share the same word stem. We apply both strategies to the target category and each anchor category and take the union of their matched labels. Instances associated with the matched target labels form the candidate target set $\mathcal{O}_{\mathrm{t}}$, while those associated with the matched anchor labels form the candidate anchor set $\mathcal{O}_{\mathrm{a}}$. 

\paragraph{Query-Relevant Voxel Construction.}
We define the query-relevant object set as $\mathcal{O}_{\mathrm{q}} = \mathcal{O}_{\mathrm{t}} \cup \mathcal{O}_{\mathrm{a}}$. Aggregating the 3D masks $m_i$ of all objects in $\mathcal{O}_{\mathrm{q}}$ yields the query-relevant voxel set, then we obtain $U = \bigcup_{o_i \in \mathcal{O}_{\mathrm{q}}} m_i$, which contains only voxels belonging to candidate target or anchor objects and therefore defines the spatial region that should receive the frame budget. QGMC uses $U$ as the coverage target and selects frames that jointly maximize its visual coverage.

\paragraph{Query-Guided Coverage Optimization.}
Finally, we select frames that jointly maximize coverage of the query-relevant voxel set $U$. Let $V_i$ denote the set of scene voxels observed in frame $f_i$. The intersection $V_i \cap U$ therefore represents the query-relevant content visible from that view. Given a frame budget $K$, QGMC selects a subset $\mathcal{E} \subseteq \mathcal{F}$ whose union covers as much of $U$ as possible:
\begin{equation*}
\max_{\mathcal{E} \subseteq \mathcal{F}} \left| \bigcup_{f_i\in\mathcal{E}} (V_i \cap U) \right|
\quad \text{s.t.} \quad |\mathcal{E}|\leq K.
\end{equation*}
The union objective favors complementary views over repeatedly observing the same query-relevant voxels. The resulting maximum coverage problem is NP-hard~\cite{feige1998threshold}, so we adopt a \emph{greedy approximation} that selects the frame with the largest marginal gain in uncovered query-relevant voxels. We show that the QGMC objective is monotone and submodular, yielding an approximation guarantee for greedy selection~\cite{nemhauser1978analysis}. Algorithm details and theoretical analysis are provided in Sec.~A of the supplementary material.

\subsection{Grounding Probability Estimation}
\label{sec:gpe}

For each candidate target $o_{t_i} \in  \mathcal{O}_{\mathrm{t}}$, we estimate a grounding probability that measures how well the object matches the description $d$. This probability combines three complementary cues: (i) the confidence score $c_{t_i}$ predicted by the instance segmentation model, (ii) the attribute consistency probabilities $\mathbb{P}_{\mathrm{a}}(o_{t_i}, p_j)$ for attributes $p_j \in \mathcal{P}$, and (iii) the relationship consistency probabilities $\mathbb{P}_{\mathrm{r}}(o_{t_i}, r_k)$ for relationships $r_k \in \mathcal{R}$. We define the grounding probability as
\begin{equation*}
\mathbb{P}(o_{t_i} | d) \propto c_{t_i}
\prod_{p_j \in \mathcal{P}} \mathbb{P}_{\mathrm{a}}(o_{t_i}, p_j)
\prod_{r_k \in \mathcal{R}} \mathbb{P}_{\mathrm{r}}(o_{t_i}, r_k),
\end{equation*}
which summarize the segmentation, attribute, and relational evidence for each candidate and are provided to the VLM as auxiliary cues for final target selection.

\paragraph{Attribute Probability Estimation.}
For each candidate target $o_{t_i}\in \mathcal{O}_{\mathrm{t}}$ and attribute $p_j\in\mathcal{P}$, we estimate how well the object's appearance matches the queried attribute. For each sampled frame $f_k\in\mathcal{E}$ in which $o_{t_i}$ is visible, we project its 3D mask $m_{t_i}$ onto the image plane and extract the corresponding crop $I_{k,t_i}$. A VLM encodes the crop and attribute into visual and text features, $\phi(I_{k,t_i})$ and $\psi(p_j)$, and computes their similarity as
\begin{equation*}
s_{t_i, j, k}^+ = \cos(\phi(I_{k, t_i}), \psi(p_j)).
\end{equation*}
To calibrate this score, we compare the same crop against a predefined reference attribute set $\mathcal{P}^-=\{p_1^-, p_2^-, \ldots, p_M^-\}$, producing
\begin{equation*}
s_{t_i, m, k}^- = \cos(\phi(I_{k,t_i}), \psi(p_m^-)).
\end{equation*}
The queried and reference similarities form the score set
\begin{equation*}
\mathcal{S}_{t_i, j, k} = \{s_{t_i, j, k}^+, s_{t_i,1,k}^-, \ldots, s_{t_i, M, k}^-\}.
\end{equation*}
We apply a quantile transformation~\cite{pedregosa2011scikit} to $\mathcal{S}_{t_i, j, k}$ to map the scores to a standard normal space, and let $z_{t_i,j,k}$ denote the transformed score corresponding to $s_{t_i, j, k}^+$. We convert it into the frame-level attribute probability
\begin{equation*}
\mathbb{P}_{\mathrm{a}}(o_{t_i}, p_j | f_k) = \Phi(z_{t_i, j, k}),
\end{equation*}
where $\Phi$ is the standard normal CDF. The final attribute probability averages over frames where $o_{t_i}$ is visible:
\begin{equation*}
\mathbb{P}_{\mathrm{a}}(o_{t_i}, p_j) = \frac{1}{|\mathcal{E}_{t_i}^{\mathrm{a}}|} \sum_{f_k \in \mathcal{E}_{t_i}^{\mathrm{a}}} \mathbb{P}_{\mathrm{a}} (o_{t_i}, p_j \mid f_k),
\end{equation*}
where $\mathcal{E}_{t_i}^{\mathrm{a}}\subseteq\mathcal{E}$ denotes the sampled frames containing $o_{t_i}$.

\begin{table*}[t]
    \caption{Quantitative comparison on ScanRefer. Results are reported as Acc@0.25 and Acc@0.5 on the \textit{Unique}, \textit{Multiple}, and \textit{Overall} splits. $\dagger$ denotes the frame-only variant without grounding probability estimation.}
    \label{tbl:scanrefer_results}
    \centering
    \setlength{\tabcolsep}{8pt}
    \resizebox{\linewidth}{!}{
    \begin{tabular}{+l ^c ^c ^c ^c ^c ^c ^c ^c}
        \toprule
        \multirow{2.5}{*}{\textbf{Method}}
        & \multirow{2.5}{*}{\textbf{Venue}}
        & \multirow{2.5}{*}{\textbf{Agent}}
        & \multicolumn{2}{c}{\textbf{Unique}}
        & \multicolumn{2}{c}{\textbf{Multiple}}
        & \multicolumn{2}{c}{\textbf{Overall}} \\
        \cmidrule(lr){4-5}\cmidrule(lr){6-7}\cmidrule(lr){8-9}
        & & & \textbf{Acc@0.25} & \textbf{Acc@0.5}
        & \textbf{Acc@0.25} & \textbf{Acc@0.5}
        & \textbf{Acc@0.25} & \textbf{Acc@0.5} \\
        \midrule
        LLM-G~\cite{yang2024llm} & ICRA 2024 & GPT-4 Turbo & - & - & - & - & 17.1 & 5.3 \\
        ZSVG3D~\cite{yuan2024visual} & CVPR 2024 & GPT-4 Turbo & 63.8 & 58.4 & 27.7 & 24.6 & 36.4 & 32.7 \\
        SeeGround~\cite{li2025seeground} & CVPR 2025 & Qwen2-VL-72B & 75.7 & 68.9 & 34.0 & 30.0 & 44.1 & 39.4 \\
        VoG~\cite{liu2026view} & AAAI 2026 & Qwen2-VL-72B & \second{78.6} & 71.5 & 34.0 & 30.4 & 44.8 & 40.3 \\
        \rowcolor{ourshl}
        \method{}$^\dagger$ (Ours) & WACV 2027 & Qwen2-VL-72B & 78.3 & \second{73.2} & \second{44.9} & \second{39.7} & \second{51.4} & \second{46.2} \\
        \rowcolor{ourshl}
        \method{} (Ours) & WACV 2027 & Qwen3-VL-8B & \first{84.9} & \first{78.7} & \first{50.2} & \first{44.3} & \first{56.9} & \first{51.0} \\
        \bottomrule
    \end{tabular}
    }
\end{table*}

\paragraph{Relationship Probability Estimation.}
For each candidate target $o_{t_i} \in \mathcal{O}_{\mathrm{t}}$ and relationship $r_j\in\mathcal{R}$, we estimate how well the target satisfies the queried spatial relation with the candidate anchor objects. For each sampled frame $f_k \in \mathcal{E}$ in which the target and at least one anchor are visible, we project their instance masks onto the image plane and annotate the target with a red bounding box and the anchors with blue bounding boxes. We then prompt a VLM to determine whether the queried relationship holds between the target and anchor objects. The VLM produces logits for a restricted set of answers (e.g., ``Yes'' and ``No''), which are normalized with a softmax; the probability assigned to ``Yes'' is taken as the frame-level relationship probability $\mathbb{P}_{\mathrm{r}}(o_{t_i}, r_j | f_k)$. We obtain the final relationship probability by averaging over all valid sampled frames:
\begin{equation*}
\mathbb{P}_{\mathrm{r}}(o_{t_i}, r_j) = \frac{1}{|\mathcal{E}_{t_i}^{\mathrm{r}}|} \sum_{f_k \in \mathcal{E}_{t_i}^{\mathrm{r}}} \mathbb{P}_{\mathrm{r}}(o_{t_i}, r_j | f_k),
\end{equation*}
where $\mathcal{E}_{t_i}^{\mathrm{r}} \subseteq \mathcal{E}$ denotes the sampled frames in which the target and at least one anchor are simultaneously visible.

\paragraph{Probability Aggregation.}
We combine the segmentation confidence, attribute probabilities, and relationship probabilities to obtain a grounding score for each candidate target $o_{t_i} \in \mathcal{O}_{\mathrm{t}}$. Assuming conditional independence among these cues, we define
\begin{equation*}
q_{t_i} = \log c_{t_i}
+ \sum_{p_j \in \mathcal{P}} \log \mathbb{P}_{\mathrm{a}}(o_{t_i}, p_j)
+ \sum_{r_k \in \mathcal{R}} \log \mathbb{P}_{\mathrm{r}}(o_{t_i}, r_k).
\end{equation*}
The log formulation improves numerical stability when combining multiple probability terms. We normalize the candidate scores with a softmax to obtain the grounding probabilities $\mathbb{P}(o_{t_i} | d) = \exp(q_{t_i}) / \sum_{o_{t_j} \in \mathcal{O}_{\mathrm{t}}}\exp(q_{t_j})$,
provided to the VLM as confidence cues for target selection.

\subsection{VLM-Based Target Selection}
\label{sec:target_selection}
Given the sampled frame set $\mathcal{E}$ and the grounding probability $\mathbb{P}(o_{t_i} | d)$ for each candidate target $o_{t_i}\in \mathcal{O}_{\mathrm{t}}$, we use a VLM to make the final grounding decision. We annotate the sampled frames with candidate object IDs and provide them to the VLM together with the textual query and estimated grounding probabilities. The VLM selects the object ID(s) that best match the query, and the corresponding 3D bounding boxes are returned as the final grounding results.

\section{Experiments}

\subsection{Experimental Setup}

\paragraph{Benchmark Datasets.}
We use two popular benchmark datasets to evaluate 3D visual grounding approaches. ScanRefer~\cite{chen2020scanrefer} contains 51,583 human-annotated referring expressions describing 11,046 objects across 800 ScanNet~\cite{dai2017scannet} scenes, with descriptions covering object attributes, spatial relationships, and contextual cues. Nr3D~\cite{achlioptas2020referit3d} contains approximately 41,503 free-form referring expressions collected on ScanNet scenes through an object reference game, emphasizing natural language variation and fine-grained disambiguation among similar object instances. These benchmarks evaluate 3D object localization from language under diverse and context-dependent referring expressions.

\paragraph{Evaluation Metrics.}
We evaluate single-object 3D visual grounding on ScanRefer~\cite{chen2020scanrefer} and Nr3D~\cite{achlioptas2020referit3d} using localization accuracy. A prediction is considered correct when the 3D Intersection over Union (IoU) between the predicted and ground-truth bounding boxes exceeds a threshold $\tau$, and the metric is denoted as Acc@$\tau$. Following the standard evaluation protocols, we report Acc@0.25 and Acc@0.5 for ScanRefer, and Acc@0.25 for Nr3D.

\paragraph{Implementation Details.}
Our framework is training-free and uses only off-the-shelf models without task-specific fine-tuning. We use Mask3D~\cite{schult2023mask3d} for 3D instance segmentation, X-VLM~\cite{zeng2022multi} to compute visual-text similarity for attribute probability estimation, and InternVL2-8B~\cite{chen2024far, chen2024internvl} for binary relationship verification. Query parsing and final grounding selection share the same VLM backbone; we evaluate Qwen2-VL-72B~\cite{wang2024qwen2} and Qwen3-VL-8B~\cite{bai2025qwen3}, using text-only prompts for query parsing. For QGMC sampling, we set the maximum frame budget to $K = 32$ and the coverage threshold to $r = 0.95$. The reference attribute set contains 617 common attributes.

\begin{figure*}
    \centering
   \includegraphics[width=\linewidth]{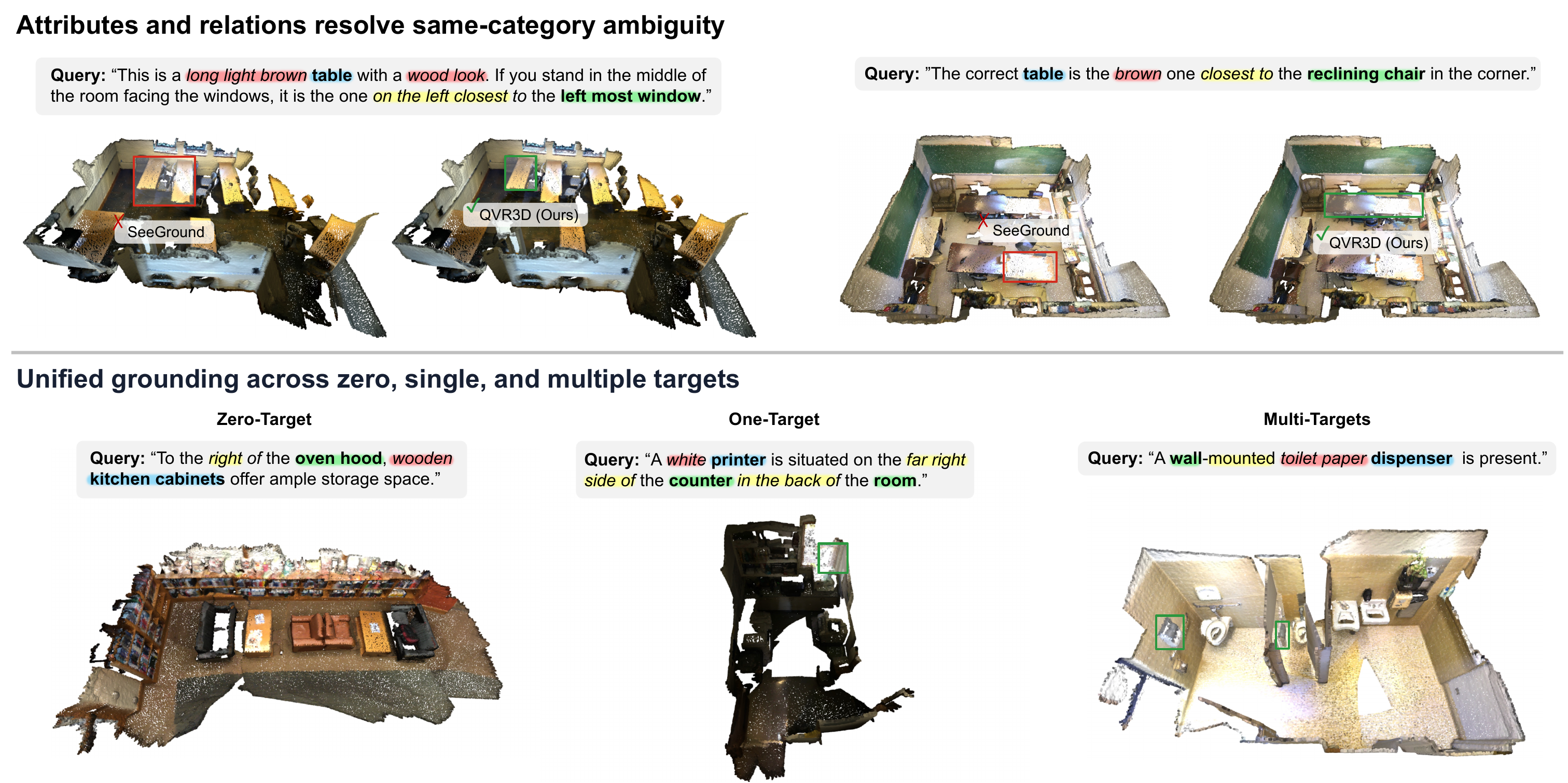}
    \caption{\textbf{Qualitative grounding results across target ambiguity and cardinality.}
Query highlights denote attributes, target categories, relationships, and anchor categories in red, blue, yellow, and green, respectively.
\textit{(Top)} On ScanRefer (left) and Nr3D (right), SeeGround selects same-category distractors (red boxes), whereas QVR3D uses the highlighted attributes and relationships to localize the intended tables (green boxes).
\textit{(Bottom)} On Multi3DRefer, QVR3D applies the same grounding pipeline to zero-, single-, and multi-target queries: from left to right, it correctly returns no object, the single matching object, and both matching objects.}
    \label{fig:qualitative}
\end{figure*}

\subsection{Quantitative Results}

Tab.~\ref{tbl:scanrefer_results} reports results on ScanRefer~\cite{chen2020scanrefer}. For Qwen2-VL-72B~\cite{wang2024qwen2}, we report the best-performing frame-only variant without grounding probability estimation, denoted by $\dagger$. This variant achieves 51.4\% Acc@0.25 and 46.2\% Acc@0.5 on the \textit{Overall} split, outperforming prior zero-shot methods using the same backbone, including SeeGround~\cite{li2025seeground} and VoG~\cite{liu2026view}. The improvement is pronounced on the \textit{Multiple} split, where grounding requires disambiguating multiple objects of the same category. Our method reaches 44.9\% Acc@0.25, demonstrating the effectiveness of query-guided frame sampling in challenging cases that require fine-grained object discrimination. Using the smaller Qwen3-VL-8B~\cite{bai2025qwen3} backbone further improves performance to 56.9\% Acc@0.25 and 51.0\% Acc@0.5 on the \textit{Overall} split, achieving the best performance among the zero-shot methods in Tab.~\ref{tbl:scanrefer_results}. The resulting performance is also competitive with several fully supervised approaches despite requiring no task-specific training.

\begin{table}[t]
    \caption{Quantitative comparison on the Nr3D benchmark. Results are reported as Acc@0.25 on the \textit{Easy}, \textit{Hard}, \textit{View-dependent}, and \textit{View-independent} splits, as well as \textit{Overall} validation set. $\dagger$ denotes the frame-only variant without grounding probability estimation.}
    \label{tbl:nr3d_results}
    \centering
    \setlength{\tabcolsep}{2pt}
    \resizebox{\linewidth}{!}{
    \begin{tabular}{l l c c c c c}
        \toprule
        \textbf{Method}
        & \textbf{Agent}
        & \textbf{Easy}
        & \textbf{Hard}
        & \textbf{Dep.}
        & \textbf{Indep.}
        & \textbf{Overall} \\
        \midrule
        ZSVG3D~\cite{yuan2024visual} & GPT-4 Turbo & 46.5 & 31.7 & 36.8 & 40.0 & 39.0 \\
        SeeGround~\cite{li2025seeground} & Qwen2-VL-72B & 54.5 & 38.3 & 42.3 & 48.2 & 46.1 \\
        LASP~\cite{mi2025language} & GPT-4o & \second{60.7} & \second{45.3} & \first{49.2} & 54.7 & \second{52.9} \\
        VoG~\cite{liu2026view} & Qwen2-VL-72B & 58.9 & 37.2 & 39.4 & 52.1 & 47.6 \\
        VIZOR~\cite{madhavaram2026vizor} & GPT-4o & \first{62.7} & 43.5 & 43.0 & \first{57.7} & 52.8 \\
        \rowcolor{ourshl}
        \method{}$^\dagger$ (Ours) & Qwen2-VL-72B & 57.4 & 43.7 & 47.0 & 52.0 & 50.3 \\
        \rowcolor{ourshl}
        \method{} (Ours) & Qwen3-VL-8B & 60.1 & \first{47.7} & \second{49.1} & \second{56.1} & \first{53.6} \\
        \bottomrule
    \end{tabular}
    }
\end{table}

Tab.~\ref{tbl:nr3d_results} reports results on Nr3D~\cite{achlioptas2020referit3d}. For Qwen2-VL-72B~\cite{wang2024qwen2}, we again report the best-performing frame-only variant without grounding probability estimation, denoted by $\dagger$. This variant outperforms SeeGround~\cite{li2025seeground} across all evaluation splits and improves over VoG~\cite{liu2026view} on the \textit{Hard}, \textit{View-dependent}, and \textit{Overall} splits. The gain on the \textit{Hard} split is particularly relevant because it contains more ambiguous expressions and visually similar objects, requiring finer object disambiguation. Using Qwen3-VL-8B~\cite{bai2025qwen3} further improves the overall performance and achieves the best zero-shot result on the \textit{Hard} split. Compared with LaSP~\cite{mi2025language}, our method performs similarly on the \textit{Easy} and \textit{View-dependent} splits while achieving higher accuracy on the \textit{Hard} and \textit{View-independent} splits, resulting in a better \textit{Overall} score. These results show that the proposed framework remains effective under challenging language and visual ambiguity without task-specific training.

We also evaluate QVR3D on Multi3DRefer~\cite{zhang2023multi3drefer} for zero-, single-, and multi-object grounding, with results reported in Supplementary Sec. B.

\subsection{Qualitative Results}
Fig.~\ref{fig:qualitative} illustrates how QVR3D resolves same-category ambiguity and handles different target cardinalities. On ScanRefer and Nr3D, SeeGround selects visually plausible but incorrect tables, whereas QVR3D identifies the intended instances by jointly considering appearance attributes and spatial relationships to anchors. The Multi3DRefer~\cite{zhang2023multi3drefer} examples show that the same inference pipeline correctly returns no object for an absent referent, one object for a single-target query, and all matching objects for a multi-target query. Together, these cases show that QVR3D uses query semantics not only to disambiguate objects but also to determine the set of referents satisfying the expression.

\subsection{Ablation Study}

\begin{table}[t]
    \caption{Ablation of framework components on ScanRefer and Nr3D using Qwen3-VL-8B. ScanRefer reports Acc@0.25 and Acc@0.5, while Nr3D reports Acc@0.25.}
    \label{tbl:ablation_qwen3vl8b}
    \resizebox{\linewidth}{!}{
    \centering
    \setlength{\tabcolsep}{5pt}
    \begin{tabular}{ccc cc c}
        \toprule
          \multirow{2.5}{*}{\textbf{QGMC}}
        & \multirow{2.5}{*}{\textbf{GPE}}
        & \multirow{2.5}{*}{\textbf{VLM-TS}}
        & \multicolumn{2}{c}{\textbf{ScanRefer}}
        & \textbf{Nr3D} \\
        \cmidrule(lr){4-5}\cmidrule(lr){6-6} & & &
        \textbf{Acc@0.25} & \textbf{Acc@0.5}
        & \textbf{Acc@0.25} \\
        \midrule
        \checkmark & \checkmark & \xmark & 44.7 & 40.3 & 36.7 \\
        \checkmark & \xmark & \checkmark & 56.1 & 50.0 & 53.1 \\
        \checkmark & \checkmark & \checkmark & \first{56.9} & \first{51.0} & \first{53.6} \\
        \bottomrule
    \end{tabular}
    }
\end{table}

\begin{table}[t]
    \caption{Ablation of frame sampling on ScanRefer with Qwen3-VL-8B under the frame-only setting. Results are reported as Acc@0.25 and Acc@0.5.}
    \resizebox{\linewidth}{!}{
    \label{tbl:ablation_sampling}
    \centering
    \begin{tabular}{lcccc}
        \toprule
        \textbf{Sampling} & \textbf{Avg. Frames} & \textbf{Acc@0.25} & \textbf{Acc@0.5} \\
        \midrule
        Uniform & 16.00 & 40.5 & 35.7 \\
        MC & 21.52 & 42.6 & 38.1 \\
        \rowcolor{ourshl}
        QGMC (Ours) & \bfseries 10.16 & \bfseries 56.1 & \bfseries 50.0 \\
        \bottomrule
    \end{tabular}
    }
\end{table}

\paragraph{Grounding Components.}
We evaluate how grounding probability estimation and VLM-based grounding selection contribute to the final prediction on ScanRefer~\cite{chen2020scanrefer} and Nr3D~\cite{achlioptas2020referit3d}. We compare three variants: \textit{probability-only}, which selects the candidate with the highest grounding probability without VLM-based selection; \textit{frame-only}, which performs VLM-based selection using QGMC-sampled frames without grounding probability estimation; and the \textit{full framework}, which combines both components. Tab.~\ref{tbl:ablation_qwen3vl8b} reports results with the Qwen3-VL-8B~\cite{bai2025qwen3} backbone. The full framework outperforms both single-component variants on ScanRefer and Nr3D. Relative to frame-only selection, adding grounding probabilities improves performance, showing that the estimated probabilities provide complementary evidence that the VLM can exploit during final grounding selection.

\paragraph{Frame Sampling.}
We compare Uniform, scene-level Maximum Coverage (MC)~\cite{zheng2025video}, and QGMC on ScanRefer. Uniform sampling uses 16 frames per video, while MC and QGMC use a maximum budget of $K=32$ with a coverage threshold of $r=0.95$. To isolate the effect of frame sampling, all variants use the Qwen3-VL-8B~\cite{bai2025qwen3} backbone under the frame-only setting without grounding probability estimation. As shown in Tab. \ref{tbl:ablation_sampling}, QGMC achieves the highest localization accuracy with the fewest frames on average. It uses 10.16 frames per query, compared with 16.00 for Uniform and 21.52 for MC, while reaching 56.1\% Acc@0.25 and 50.0\% Acc@0.5, versus 42.6\% and 38.1\% for MC. These results show that query-relevant frame allocation improves both grounding accuracy and sampling efficiency over scene-level coverage.

\begin{table}[t]
    \caption{Comparison of different frame sampling strategies for Video-3D LLM on the Multi3DRefer benchmark. Results are reported as F1 score at IoU thresholds of 0.25 (F1@0.25) and 0.5 (F1@0.5).}
    \label{tbl:sampling_multi3drefer}
    \centering
    \resizebox{\linewidth}{!}{
    \begin{tabular}{cccccc}
        \toprule
        \textbf{Frame} & \multirow{2}{*}{\textbf{Sampling}} & \textbf{Coverage} & \textbf{Avg.} & \multirow{2}{*}{\textbf{F1@0.25}} & \multirow{2}{*}{\textbf{F1@0.5}} \\
        \textbf{Budget} & ~ & \textbf{Threshold} & \textbf{Frames} & ~ & ~ \\
        \midrule
        8 & Uniform & - & 8.00 & 49.80 & 45.40 \\
        8 & MC & 1.0 & 8.00 & 53.55 & 48.54 \\
        16 & Uniform & - & 16.00 & 54.95 & 49.82 \\
        16 & MC & 1.0 & 16.00 & 56.65 & 51.39 \\
        32 & Uniform & - & 31.99 & 58.02 & 52.68 \\
        32 & MC & 1.0 & 31.77 & 57.93 & 52.50 \\
        32 & MC & 0.95 & 21.33 & 57.87 & 52.40 \\
        \rowcolor{ourshl}
        32 & QGMC & 0.95 & 10.53 & \first{58.09} & \first{53.03} \\
        \bottomrule
    \end{tabular}
    }
\end{table}

\subsection{Transferability of QGMC Sampling}
We evaluate the transferability of QGMC by integrating it into Video-3D LLM~\cite{zheng2025video}, a frame-based 3D vision-language model for visual grounding, dense captioning, and question answering. We replace its frame sampler with QGMC while keeping all other components unchanged. This isolates the effect of frame selection and evaluates QGMC as a plug-and-play component without retraining or task-specific adaptation. Tab.~\ref{tbl:sampling_multi3drefer} reports results on Multi3DRefer~\cite{zhang2023multi3drefer}. QGMC achieves the best F1@0.25 and F1@0.5 among the evaluated sampling strategies while using only 10.53 frames on average. This corresponds to less than half the frames required by Maximum Coverage (MC) with a coverage threshold of 0.95 and approximately one-third of the 32-frame budget, demonstrating that query-guided sampling can improve grounding accuracy while substantially reducing visual input. Additional results across 3D vision-language tasks are provided in Sec. B of the supplementary material.

\section{Conclusion}
We introduce QVR3D, a query-guided visual reasoning framework for training-free 3D grounding. By combining QGMC for query-relevant view selection with GPE for integrating instance, attribute, and relational evidence, QVR3D supports both single- and multi-object grounding without task-specific adaptation. Experiments on ScanRefer and Nr3D demonstrate state-of-the-art zero-shot performance. QGMC generalizes as a plug-and-play sampler across frame-based 3D vision-language tasks, improving frame efficiency.

\section*{Acknowledgements}

This work was supported by the National Science and Technology Council (NSTC), Taiwan, under Grants NSTC 114-2221-E-002-182-MY3, 114-2634-F-002-004, 115-2218-E-001-003, 115-2628-E-001-001-MY2, 115-2634-F-001-007, 115-2628-E-A49-007, and 115-2634-F-A49-011, and by Academia Sinica under Grant AS-IAIA-114-M10. We thank the National Center for High-performance Computing (NCHC) of the National Applied Research Laboratories (NARLabs), Taiwan, for providing computational and storage resources. We acknowledge the generous support of the National Science and Technology Center for Disaster Reduction (NCDR), and the Industrial Technology Research Institute (ITRI), Taiwan. Yu-Lun Liu acknowledges support from the Yushan Young Fellow Program of the Ministry of Education (MOE), Taiwan.

{
    \small
    \bibliographystyle{ieeenat_fullname}
    \bibliography{main}
}

\clearpage
\renewcommand\thesection{\Alph{section}}
\setcounter{section}{0}

\twocolumn[
    \vspace{1cm}
    \begin{center}
        \Large{\textbf{[Supplementary Material] QVR3D: Query-Guided Visual Reasoning for Training-Free 3D Grounding}}
    \end{center}
    \vspace{1cm}
]

\DoToC
\section{Algorithmic and Theoretical Details of QGMC}

\subsection{Greedy Optimization}
We solve the QGMC objective with a greedy approximation. Let $\mathcal{E}^{(k)}$ denote the set of frames selected after $k$ iterations and $T^{(k)}$ the query-relevant voxels covered by these frames, with $\mathcal{E}^{(0)}=\emptyset$ and $T^{(0)}=\emptyset$. At iteration $k+1$, QGMC selects the frame with the largest marginal gain in uncovered query-relevant voxels:
\begin{equation*}
f^{(k+1)} = \argmax_{f_i \in \mathcal{F}\setminus\mathcal{E}^{(k)}}
\left|(V_i\cap U)\setminus T^{(k)}\right|.
\end{equation*}
The selected frame set and covered voxel set are then updated recursively as
\begin{equation*}
\mathcal{E}^{(k+1)} = \mathcal{E}^{(k)} \cup \left\{f^{(k+1)}\right\},
\end{equation*}
and
\begin{equation*}
T^{(k+1)} = T^{(k)} \cup \left(V^{(k+1)}\cap U\right),
\end{equation*}
where $V^{(k+1)}$ denotes the voxels observed by $f^{(k+1)}$. The procedure terminates when $|\mathcal{E}^{(k)}|=K$ or the coverage ratio $|T^{(k)}|/|U|$ reaches the threshold $r$. The complete procedure is summarized in Algo. \ref{algo:qgmc}. The QGMC objective is a variant of the maximum coverage problem, whose theoretical properties are analyzed in the following subsection.

\begin{algorithm}[t]
\caption{QGMC Sampling}
\label{algo:qgmc}
\begin{algorithmic}[1]
\Require Set of video frames $\mathcal{F} = \left\{f_1, f_2, \ldots, f_n\right\}$, voxel set $V_i$ of frame $f_i$, query-relevant voxel set $U$, frame budget $K$, coverage threshold $r$.
\Ensure Frame subset $\mathcal{E} \subseteq \mathcal{F}$ maximizing coverage of query-relevant voxels.

\State $\mathcal{E} \gets \varnothing$
\State $T \gets \varnothing$ \Comment{Covered query-relevant voxels}

\While{$\left\lvert \mathcal{E} \right\rvert < K$}
    \State $f^* \gets \argmax_{f_i \in \mathcal{F} \setminus \mathcal{E}}\left\lvert \left(V_i \cap U\right) \setminus T \right\rvert$
    \State $\mathcal{E} \gets \mathcal{E} \cup \{f^*\}$
    \State $T \gets T \cup (V^* \cap U)$ \Comment{$V^*$ denotes the voxel set of frame $f^*$}
    
    \If{$|T| / |U| \geq r$}
        \State \textbf{break}
    \EndIf
\EndWhile

\State \Return $\mathcal{E}$
\end{algorithmic}
\end{algorithm}

\subsection{Theoretical Analysis}
We analyze QGMC through the classical maximum coverage formulation. The coverage objective is monotone and submodular, which yields the standard approximation guarantee for greedy selection. Let
\begin{equation*}
g(\mathcal{E}) := \left| \bigcup_{f_i \in \mathcal{E}}(V_i \cap U) \right|
\end{equation*}
denote the number of query-relevant voxels covered by a selected frame set $\mathcal{E}$.

\begin{lemma}
\label{lemma:monotonicity}
The objective function $g(\mathcal{E})$ is monotone non-decreasing.
\end{lemma}

\begin{proof}
Consider two frame subsets $\mathcal{E}_1 \subseteq \mathcal{E}_2 \subseteq \mathcal{F}$. Since every frame contained in $\mathcal{E}_1$ is also contained in $\mathcal{E}_2$, we have
\begin{equation*}
\bigcup_{f_i \in \mathcal{E}_1} (V_i \cap U) \subseteq \bigcup_{f_i \in \mathcal{E}_2}(V_i \cap U).
\end{equation*}
Therefore, $g(\mathcal{E}_1) \leq g(\mathcal{E}_2)$.
\end{proof}

\begin{lemma}
\label{lemma:submodularity}
The objective function $g(\mathcal{E})$ is submodular.
\end{lemma}

\begin{proof}
Let $\mathcal{E}_1 \subseteq \mathcal{E}_2 \subseteq \mathcal{F}$ and let $f_x \in \mathcal{F} \setminus \mathcal{E}_2$ be an unselected frame. The marginal gain of adding frame $f_x$ to a frame subset $\mathcal{E}$ is
\begin{equation*}
\Delta(f_x \mid \mathcal{E}) = g(\mathcal{E} \cup \{f_x\}) - g(\mathcal{E}).
\end{equation*}
Denote the query-relevant voxels already covered by $\mathcal{E}$ as
\begin{equation*}
C_\mathcal{E} = \bigcup_{f_i \in \mathcal{E}} (V_i \cap U).
\end{equation*}
The marginal gain can then be rewritten as
\begin{equation*}
\Delta(f_x \mid \mathcal{E}) = |(V_x \cap U) \setminus C_\mathcal{E}|.
\end{equation*}
Since $\mathcal{E}_1 \subseteq \mathcal{E}_2$, it follows that $C_{\mathcal{E}_1} \subseteq C_{\mathcal{E}_2}.$
Therefore,
\begin{equation*}
(V_x \cap U) \setminus C_{\mathcal{E}_1} \supseteq (V_x \cap U) \setminus C_{\mathcal{E}_2},
\end{equation*}
which implies
\begin{equation*}
\Delta(f_x \mid \mathcal{E}_1) \ge \Delta(f_x \mid \mathcal{E}_2).
\end{equation*}

The marginal gain obtained by adding a frame never increases as more frames are selected, satisfying the diminishing returns property. Hence $g(\mathcal{E})$ is submodular.
\end{proof}

\begin{theorem}[Greedy Approximation Guarantee]
Let $\mathcal{E}_{\mathrm{greedy}}$ denote the frame set obtained by running greedy QGMC for $K$ iterations, and let $\mathcal{E}_{\mathrm{opt}} \in \argmax_{|\mathcal{E}|\leq K} g(\mathcal{E})$ be an optimal solution. Then,
\begin{equation*}
g(\mathcal{E}_{\mathrm{greedy}}) \geq \left(1 - \frac{1}{\mathrm{e}}\right) g(\mathcal{E}_{\mathrm{opt}}),
\end{equation*}
where $\mathrm{e}$ denotes Euler's number.
\end{theorem}

\begin{proof}
Let $\mathcal{E}^{(k)}$ denote the frame set selected by the greedy algorithm after $k$ iterations, with $\mathcal{E}^{(0)} = \varnothing$. Since $g$ is monotone and submodular by Lemma~\ref{lemma:monotonicity} and Lemma~\ref{lemma:submodularity}, and $g(\varnothing) = 0$, we have
\begin{equation*}
\begin{aligned}
g(\mathcal{E}_{\mathrm{opt}}) &\leq g\left(\mathcal{E}^{(k)}\cup\mathcal{E}_{\mathrm{opt}}\right) \\
&\leq g(\mathcal{E}^{(k)}) + \sum_{f\in\mathcal{E}_{\mathrm{opt}}\setminus\mathcal{E}^{(k)}}
\left[g\left(\mathcal{E}^{(k)}\cup\{f\}\right) - g(\mathcal{E}^{(k)}) \right].
\end{aligned}
\end{equation*}
Because $|\mathcal{E}_{\mathrm{opt}}|\leq K$, at least one frame
$f \in \mathcal{E}_{\mathrm{opt}}\setminus\mathcal{E}^{(k)}$
has marginal gain satisfying
\begin{equation*}
g(\mathcal{E}^{(k)}\cup\{f\}) - g(\mathcal{E}^{(k)})
\geq \frac{1}{K} (g(\mathcal{E}_{\mathrm{opt}}) - g(\mathcal{E}^{(k)})).
\end{equation*}
Since greedy QGMC selects the frame with the largest marginal gain,
\begin{equation*}
g(\mathcal{E}^{(k+1)}) - g(\mathcal{E}^{(k)}) \geq
\frac{1}{K}
(g(\mathcal{E}_{\mathrm{opt}}) - g(\mathcal{E}^{(k)})).
\end{equation*}
Hence
\begin{equation*}
g(\mathcal{E}_{\mathrm{opt}})
- g(\mathcal{E}^{(k+1)}) \leq
\left(1-\frac{1}{K}\right) (g(\mathcal{E}_{\mathrm{opt}}) - g(\mathcal{E}^{(k)})).
\end{equation*}
Applying this inequality recursively gives
\begin{equation*}
g(\mathcal{E}_{\mathrm{opt}})
- g(\mathcal{E}^{(k)}) \leq
\left(1-\frac{1}{K}\right)^k (g(\mathcal{E}_{\mathrm{opt}}) - g(\mathcal{E}^{(0)})).
\end{equation*}
Since $\mathcal{E}^{(0)}=\varnothing$ and $g(\varnothing) = 0$,
\begin{equation*}
g(\mathcal{E}^{(k)})
\geq \left[1- \left(1-\frac{1}{K}\right)^k\right]
g(\mathcal{E}_{\mathrm{opt}}).
\end{equation*}
Setting $k=K$ yields
\begin{equation*}
g(\mathcal{E}_{\mathrm{greedy}})
\geq \left[1 - \left(1-\frac{1}{K}\right)^K \right]
g(\mathcal{E}_{\mathrm{opt}}).
\end{equation*}
Finally, using
\begin{equation*}
\left(1-\frac{1}{K}\right)^K
\leq
\frac{1}{\mathrm{e}},
\end{equation*}
we obtain
\begin{equation*}
g(\mathcal{E}_{\mathrm{greedy}})
\geq
\left(1-\frac{1}{\mathrm{e}}\right)
g(\mathcal{E}_{\mathrm{opt}}).
\end{equation*}
\end{proof}

Thus, although exact QGMC optimization is NP-hard, greedy selection provides an efficient solution with a constant-factor approximation guarantee.

\begin{table*}[t]
    \caption{Quantitative comparison on the Multi3DRefer benchmark. We report F1 score (F1) under IoU thresholds of 0.25 and 0.5 on the \textit{Zero Target (ZT)}, \textit{Single Target (ST)}, and \textit{Multiple Target (MT)} splits, with and without distractors (\textit{w/ D} and \textit{w/o D}), as well as \textit{Overall} validation set. Methods are grouped into supervised expert, supervised generalist, and zero-shot settings.}
    \label{tbl:multi3drefer_results}
    \centering
    \setlength{\tabcolsep}{6pt}
    \resizebox{\linewidth}{!}{
    \begin{tabular}{l cc cccc cc cc}
        \toprule
        \multirow{4}{*}{\textbf{Method}}
        & \multicolumn{2}{c}{\textbf{ZT}}
        & \multicolumn{4}{c}{\textbf{ST}}
        & \multicolumn{2}{c}{\multirow{2}{*}{\textbf{MT}}}
        & \multicolumn{2}{c}{\multirow{2}{*}{\textbf{Overall}}} \\
        \cmidrule(lr){2-3}\cmidrule(lr){4-7}
        &
        \textbf{w/o D} & \textbf{w/ D}
        & \multicolumn{2}{c}{\textbf{w/o D}} & \multicolumn{2}{c}{\textbf{w/ D}}
        &&&& \\
        \cmidrule(lr){4-5}\cmidrule(lr){6-7}
        &
        \textbf{F1} & \textbf{F1} & \textbf{F1@0.25} & \textbf{F1@0.5}
        & \textbf{F1@0.25} & \textbf{F1@0.5}
        & \textbf{F1@0.25} & \textbf{F1@0.5}
        & \textbf{F1@0.25} & \textbf{F1@0.5} \\
        \midrule
        \multicolumn{11}{l}{\textbf{\textit{Supervised Expert}}} \\
        M3DRef-CLIP~\cite{zhang2023multi3drefer} & 81.8 & 39.4 & 53.5 & 47.8 & 34.6 & 30.6 & 43.6 & 37.9 & 42.8 & 38.4 \\
        \midrule
        \multicolumn{11}{l}{\textbf{\textit{Supervised Generalist}}} \\
        D3Net~\cite{chen2022d} & 81.6 & 32.5 & - & 38.6 & - & 23.3 & - & 35.0 & - & 32.2 \\
        3DJCG~\cite{cai20223djcg} & 94.1 & 66.9 & - & 26.0 & - & 16.7 & - & 26.2 & - & 26.6 \\
        Grounded 3D-LLM~\cite{chen2024grounded} & - & - & - & - & - & - & - & - & 45.2 & 40.6 \\
        PQ3D~\cite{zhu2024unifying} & 85.4 & 57.7 & - & 68.5 & - & 43.6 & - & 40.9 & - & 50.1 \\
        ChatScene~\cite{huang2024chat} & 90.3 & 62.6 & 82.9 & 75.9 & 49.1 & 44.5 & 45.7 & 41.1 & 57.1 & 52.4 \\
        Video-3D LLM~\cite{zheng2025video} & 94.7 & 78.5 & 82.6 & 73.4 & 52.1 & 47.2 & 40.8 & 35.7 & 58.0 & 52.7 \\
        \midrule
        \multicolumn{11}{l}{\textbf{\textit{Zero-Shot}}} \\
        \rowcolor{ourshl}
        \method{} (Ours) & 93.8 & 76.7 & 69.8 & 64.9 & 48.5 & 44.0 & 45.6 & 40.7 & 54.9 & 50.6 \\
        \bottomrule
    \end{tabular}
    }
\end{table*}

\section{More Experiments}
\label{sec:more_exp}

\subsection{Implementation Details}
Experiments using Qwen2-VL-72B are conducted on two NVIDIA H100 GPUs, while those using Qwen3-VL-8B are conducted on a single NVIDIA H100 GPU.

\subsection{Benchmark Settings}

\paragraph{ScanRefer.}
ScanRefer~\cite{chen2020scanrefer} evaluates single-object 3D visual grounding using localization accuracy at IoU thresholds of 0.25 and 0.5, reported as Acc@0.25 and Acc@0.5. The \textit{Unique} split contains scenes with a single instance of the target category, while the \textit{Multiple} split contains multiple same-category instances and therefore requires finer object disambiguation.

\paragraph{Nr3D.}
Nr3D~\cite{achlioptas2020referit3d} evaluates localization using Acc@0.25. The benchmark is divided into \textit{Easy} and \textit{Hard} splits according to referring-expression ambiguity, and into \textit{View-dependent} and \textit{View-independent} splits according to whether the description depends on observer's viewpoint.

\paragraph{Multi3DRefer.}
Multi3DRefer~\cite{zhang2023multi3drefer} extends 3D visual grounding to queries referring to zero, one, or multiple objects. Performance is measured using F1@0.25 and F1@0.5, which jointly account for precision and recall under the corresponding IoU thresholds. The benchmark includes \textit{Zero Target (ZT)}, \textit{Single Target (ST)}, and \textit{Multiple Target (MT)} settings, with the latter two further divided according to the presence of same-category distractors (\textit{w/ D} and \textit{w/o D}). For ZT queries, the F1 score is 1 when the model correctly predicts no target and 0 otherwise.

\subsection{Multi3DRefer Results}
Table~\ref{tbl:multi3drefer_results} reports results on Multi3DRefer~\cite{zhang2023multi3drefer}. To the best of our knowledge, QVR3D is the first zero-shot method evaluated on this benchmark, extending training-free 3D grounding from single-target localization to queries with multiple valid targets. Using Qwen3-VL-8B~\cite{bai2025qwen3}, our method achieves 54.9\% F1@0.25 and 50.6\% F1@0.5 on the \textit{Overall} split, remaining competitive with supervised generalist models such as PQ3D~\cite{zhu2024unifying}, ChatScene~\cite{huang2024chat}, and Video-3D LLM~\cite{zheng2025video}. On the more challenging \textit{Multiple Target (MT)} split, QVR3D reaches 45.6\% F1@0.25 and 40.7\% F1@0.5, demonstrating its ability to localize multiple target instances without task-specific training or architectural adaptation.

\begin{table}[t]
    \caption{Ablation of framework components on ScanRefer and Nr3D using Qwen2-VL-72B. ScanRefer reports Acc@0.25 and Acc@0.5, while Nr3D reports Acc@0.25.}
    \label{tbl:ablation_qwen2vl72b}
    \centering
    \resizebox{\linewidth}{!}{
    \begin{tabular}{ccc cc c}
        \toprule
        \multirow{2}{*}{\textbf{QGMC}}
        & \multirow{2}{*}{\textbf{GPE}}
        & \multirow{2}{*}{\textbf{VLM-TS}}
        & \multicolumn{2}{c}{\textbf{ScanRefer}}
        & \textbf{Nr3D} \\
        \cmidrule(lr){4-5}\cmidrule(lr){6-6}
        & & & \textbf{Acc@0.25} & \textbf{Acc@0.5}
        & \textbf{Acc@0.25} \\
        \midrule
         \checkmark & \checkmark & \xmark & 44.2 & 40.0 & 35.6 \\
        \checkmark & \xmark & \checkmark & \bfseries 51.4 & \bfseries 46.2 & \bfseries 50.3 \\
        \checkmark & \checkmark & \checkmark & 49.5 & 44.6 & 46.4 \\
        \bottomrule
    \end{tabular}
    }
\end{table}

\subsection{Ablation of Grounding Components on Qwen2-VL-72B}
Tab.~\ref{tbl:ablation_qwen2vl72b} reports the same ablation with Qwen2-VL-72B~\cite{wang2024qwen2}. The frame-only variant achieves the best performance, while adding grounding probabilities consistently degrades the results. In contrast to Qwen3-VL-8B, which benefits from GPE, Qwen2-VL-72B appears less effective at exploiting numerical confidence cues during target selection. This indicates that the benefit of GPE is backbone-dependent and may depend on the VLM's ability to interpret and utilize numerical probability information.

\subsection{Qualitative Results}

\begin{figure*}[t]
\centering
\includegraphics[width=\textwidth]{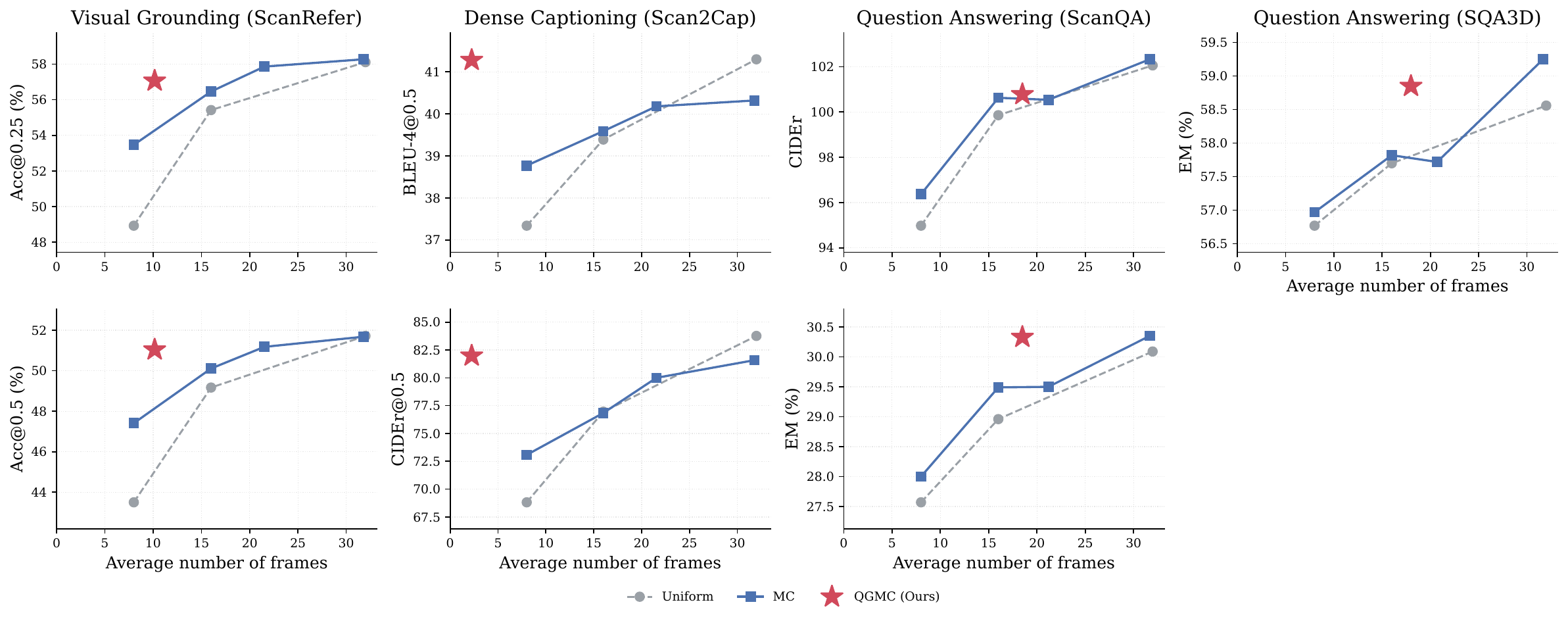}    \caption{\textbf{Frame efficiency of query-guided sampling across four benchmarks.} Each column corresponds to one benchmark; when two metrics are reported, the primary metric is shown above and the secondary metric below. The horizontal axis denotes the average number of frames processed by the model. Uniform and MC are evaluated with budgets of 8, 16, and 32 frames. Across all metrics, QGMC achieves a stronger performance-efficiency trade-off at its operating point while requiring only 2.25-18.49 frames on average.}
\label{fig:pareto}
\end{figure*}

\begin{figure*}[t]
    \centering
    \includegraphics[width=\textwidth]{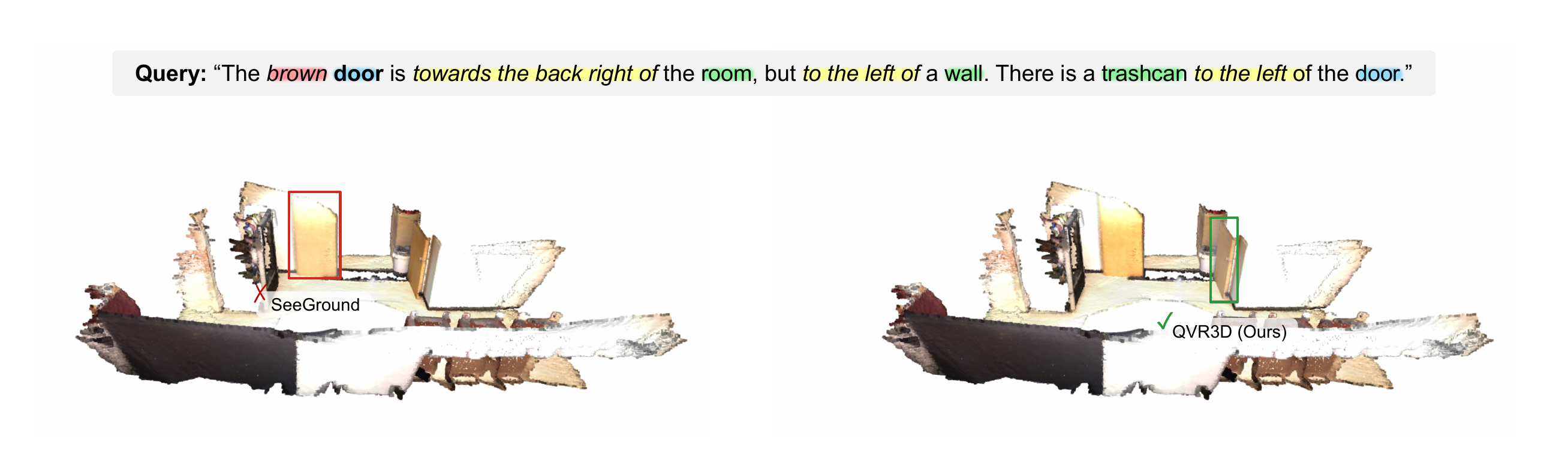}
    \caption{Additional qualitative result on ScanRefer. The query identifies the target door through its spatial relations to the wall and trash can. SeeGround selects a distractor (red box), whereas QVR3D localizes the relation-consistent door (green box).}
    \label{fig:supp_qual_scanrefer_door}
\end{figure*}

\begin{figure*}[t]
    \centering
    \includegraphics[width=\textwidth]{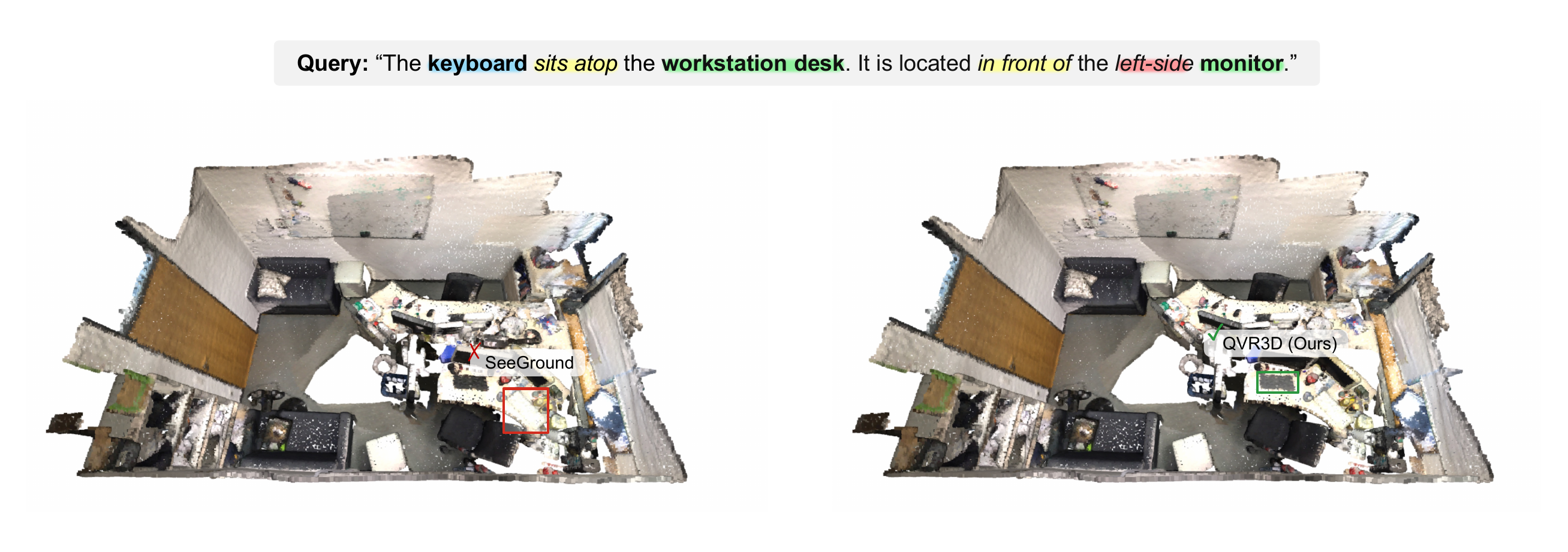}
    \caption{Additional qualitative result on ScanRefer. The keyboard is identified through its relations to the workstation desk and the left-side monitor. SeeGround selects a distractor (red box), whereas QVR3D localizes the relation-consistent keyboard (green box).}
    \label{fig:supp_qual_scanrefer_keyboard}
\end{figure*}

\paragraph{ScanRefer.}
We further compare qualitative grounding results on the ScanRefer~\cite{chen2020scanrefer} benchmark using the Qwen2-VL-72B~\cite{wang2024qwen2} backbone. Figures~\ref{fig:supp_qual_scanrefer_door} and~\ref{fig:supp_qual_scanrefer_keyboard} present representative examples comparing our method with SeeGround~\cite{li2025seeground}. In all cases, both methods correctly identify the target category, but SeeGround fails to distinguish the referred object from other same-category instances in the scene. This is particularly evident when the query relies on spatial relationships, such as object proximity, relative position, or surrounding context. As a result, SeeGround often selects a semantically similar but incorrect object, whereas our method successfully localizes the ground-truth target. These examples highlight the challenges of object disambiguation and show that our method can more reliably identify the referred object in such scenarios.

\begin{figure*}[t]
    \centering
    \includegraphics[width=\textwidth]{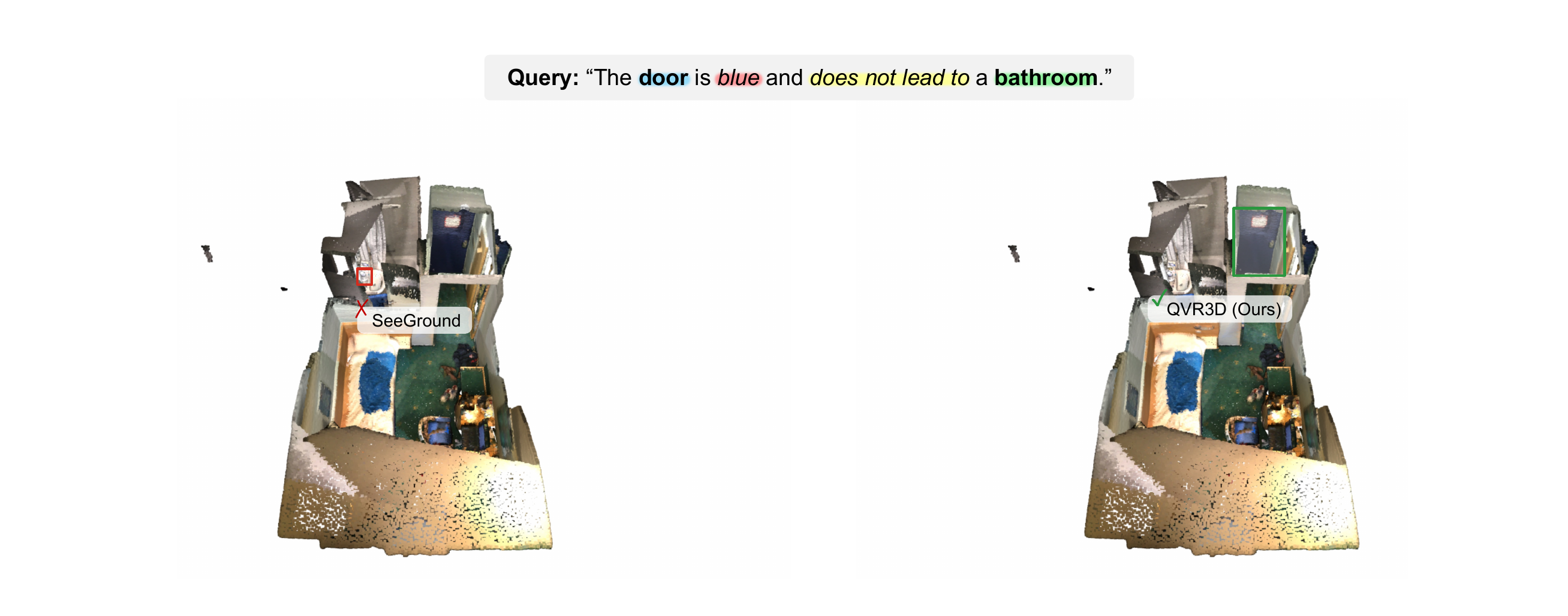}
    \caption{Additional qualitative result on Nr3D. The negative contextual constraint excludes the bathroom door. SeeGround selects a distractor (red box), whereas QVR3D localizes the described blue door (green box).}
    \label{fig:supp_qual_nr3d_door}
\end{figure*}

\begin{figure*}[t]
    \centering
    \includegraphics[width=\textwidth]{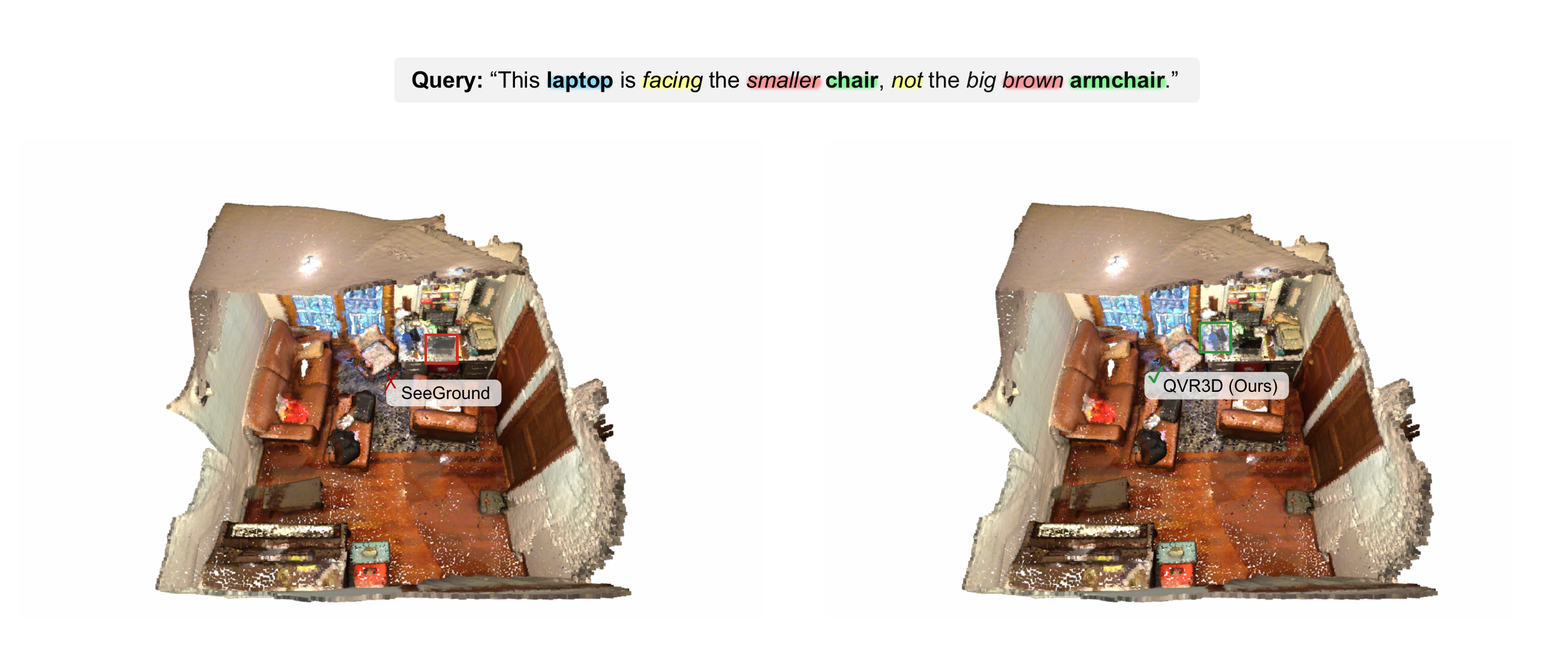}
    \caption{Additional qualitative result on Nr3D. The contrastive relation to the smaller chair, rather than the large brown armchair, disambiguates the intended laptop. SeeGround selects a distractor (red box), whereas QVR3D localizes the relation-consistent laptop (green box).}
    \label{fig:supp_qual_nr3d_laptop}
\end{figure*}

\begin{figure*}[t]
    \centering
    \includegraphics[width=\textwidth]{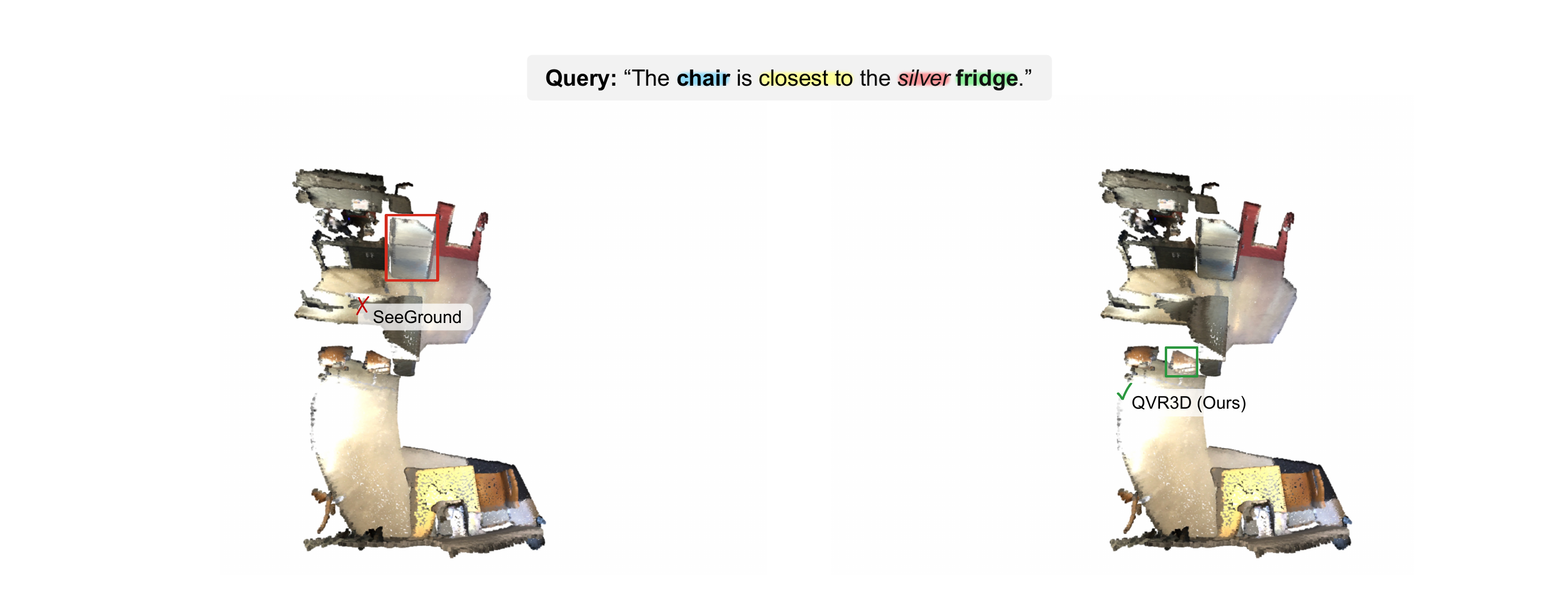}
    \caption{Additional qualitative result on Nr3D. Proximity to the silver fridge disambiguates the target chair. SeeGround selects a distractor (red box), whereas QVR3D localizes the intended instance (green box).}
    \label{fig:supp_qual_nr3d_chair}
\end{figure*}

\paragraph{Nr3D.}
We also compare qualitative grounding results on the Nr3D~\cite{achlioptas2020referit3d} benchmark using the Qwen2-VL-72B~\cite{wang2024qwen2} backbone. Figures~\ref{fig:supp_qual_nr3d_door}, \ref{fig:supp_qual_nr3d_laptop}, and~\ref{fig:supp_qual_nr3d_chair} present representative examples comparing our method with SeeGround~\cite{li2025seeground}. As shown in the figures, SeeGround exhibits two common failure modes. In some cases, it fails to correctly identify the target category described in the query, resulting in predictions on semantically unrelated objects. In other cases, although the target category is correctly recognized, it fails to distinguish the referred object from other instances of the same category in the scene. These errors are particularly evident when the query requires reasoning about object attributes, relative positions, or surrounding contextual cues. In contrast, our method successfully identifies the target object in all presented examples. These qualitative results suggest that the proposed framework can better utilize query-relevant visual evidence and contextual information, resulting in more reliable object categorization and object disambiguation in challenging grounding scenarios.

\subsection{Generalization of QGMC Across 3D Vision-Language Tasks}

We evaluate the generality of QGMC by replacing the original frame sampler in Video-3D LLM~\cite{zheng2025video} while keeping its architecture, training, and inference pipeline unchanged. This isolates the effect of frame selection and evaluates QGMC as a plug-and-play component across visual grounding, dense captioning, and question answering. Because different sampling strategies yield different numbers of input frames, we evaluate baselines across multiple frame budgets and compare task performance against the average number of frames actually processed by the model.

As shown in Fig.~\ref{fig:pareto}, QGMC consistently achieves a stronger performance-efficiency trade-off across benchmarks, recovering the performance of substantially larger frame budgets with far fewer observations. The gain therefore stems from allocating limited visual capacity to query-relevant evidence rather than processing more input. This advantage is largest for dense captioning, where the referent is spatially localized, and smaller for question answering, which often requires broader scene context beyond target and anchor regions. These results indicate that the visual budget required for 3D vision-language reasoning is more closely tied to the spatial extent of query-relevant evidence than to overall scene size.

\begin{table}[t]
    \caption{\textbf{Comparison of different frame sampling strategies for visual grounding on ScanRefer}. Results are reported as localization accuracy at IoU thresholds of 0.25 and 0.5.}
    \label{tbl:sampling_scanrefer}
    \centering
    \resizebox{\linewidth}{!}{
    \begin{tabular}{+c ^c ^c ^c ^c ^c}
        \toprule
        \textbf{Frame} & \multirow{2}{*}{\textbf{Sampling}} & \textbf{Coverage} & \textbf{Avg.} & \multirow{2}{*}{\textbf{Acc@0.25}} & \multirow{2}{*}{\textbf{Acc@0.5}} \\
        \textbf{Budget} & ~ & \textbf{Threshold} & \textbf{Frames} & ~ & ~ \\
        \midrule
        8 & Uniform & - & 8.00 & 48.93 & 43.50 \\
        8 & MC & 1.0 & 8.00 & 53.47 & 47.41 \\
        \midrule
        16 & Uniform & - & 16.00 & 55.42 & 49.17 \\
        16 & MC & 1.0 & 16.00 & 56.46 & 50.11 \\
        \midrule
        32 & Uniform & - & 31.99 & 58.11 & 51.72 \\
        32 & MC & 1.0 & 31.78 & 58.27 & 51.68 \\
        32 & MC & 0.95 & 21.52 & 57.86 & 51.18 \\
        \rowcolor{ourshl}
        32 & QGMC & 0.95 & 10.16 & 57.06 & 51.04 \\
        \bottomrule
    \end{tabular}
}
\end{table}

\begin{table}[t]
    \caption{\textbf{Comparison of different frame sampling strategies for dense captioning on Scan2Cap}. Results are reported as BLEU-4 and CIDEr at an IoU threshold of 0.5.}
    \label{tbl:sampling_scan2cap}
    \centering
    \resizebox{\linewidth}{!}{
    \begin{tabular}{+c ^c ^c ^c ^c ^c}
        \toprule
        \textbf{Frame} & \multirow{2}{*}{\textbf{Sampling}} & \textbf{Coverage} & \textbf{Avg.} & \multirow{2}{*}{\textbf{BLEU-4@0.5}} & \multirow{2}{*}{\textbf{CIDEr@0.5}} \\
        \textbf{Budget} & ~ & \textbf{Threshold} & \textbf{Frames} & ~ & ~ \\
        \midrule
        8 & Uniform & - & 8.00 & 37.34 & 68.82 \\
        8 & MC & 1.0 & 8.00 & 38.77 & 73.08 \\
        \midrule
        16 & Uniform & - & 16.00 & 39.39 & 76.96 \\
        16 & MC & 1.0 & 16.00 & 39.59 & 76.84 \\
        \midrule
        32 & Uniform & - & 31.99 & 41.30 & 83.76 \\
        32 & MC & 1.0 & 31.78 & 40.32 & 81.58 \\
        32 & MC & 0.95 & 21.55 & 40.18 & 80.00 \\
        \rowcolor{ourshl}
        32 & QGMC & 0.95 & 2.25 & 41.28 & 81.98 \\
        \bottomrule
    \end{tabular}
}
\end{table}

\paragraph{Visual Grounding.}
Tab.~\ref{tbl:sampling_scanrefer} reports results on ScanRefer~\cite{chen2020scanrefer}. Compared with 16-frame MC, QGMC improves both Acc@0.25 and Acc@0.5 while reducing the average input from 16.00 to 10.16 frames. These results show that prioritizing query-relevant regions improves both grounding accuracy and frame efficiency.

\paragraph{Dense Captioning.}
Tab.~\ref{tbl:sampling_scan2cap} reports dense captioning results on Scan2Cap~\cite{chen2021scan2cap} using BLEU-4@0.5~\cite{papineni2002bleu} and CIDEr@0.5~\cite{vedantam2015cider}. QGMC outperforms MC while using only 2.25 frames on average, indicating that focused observations of query-relevant objects are sufficient for accurate localization and description.

\begin{figure*}[t]
\centering
\includegraphics[width=\textwidth]{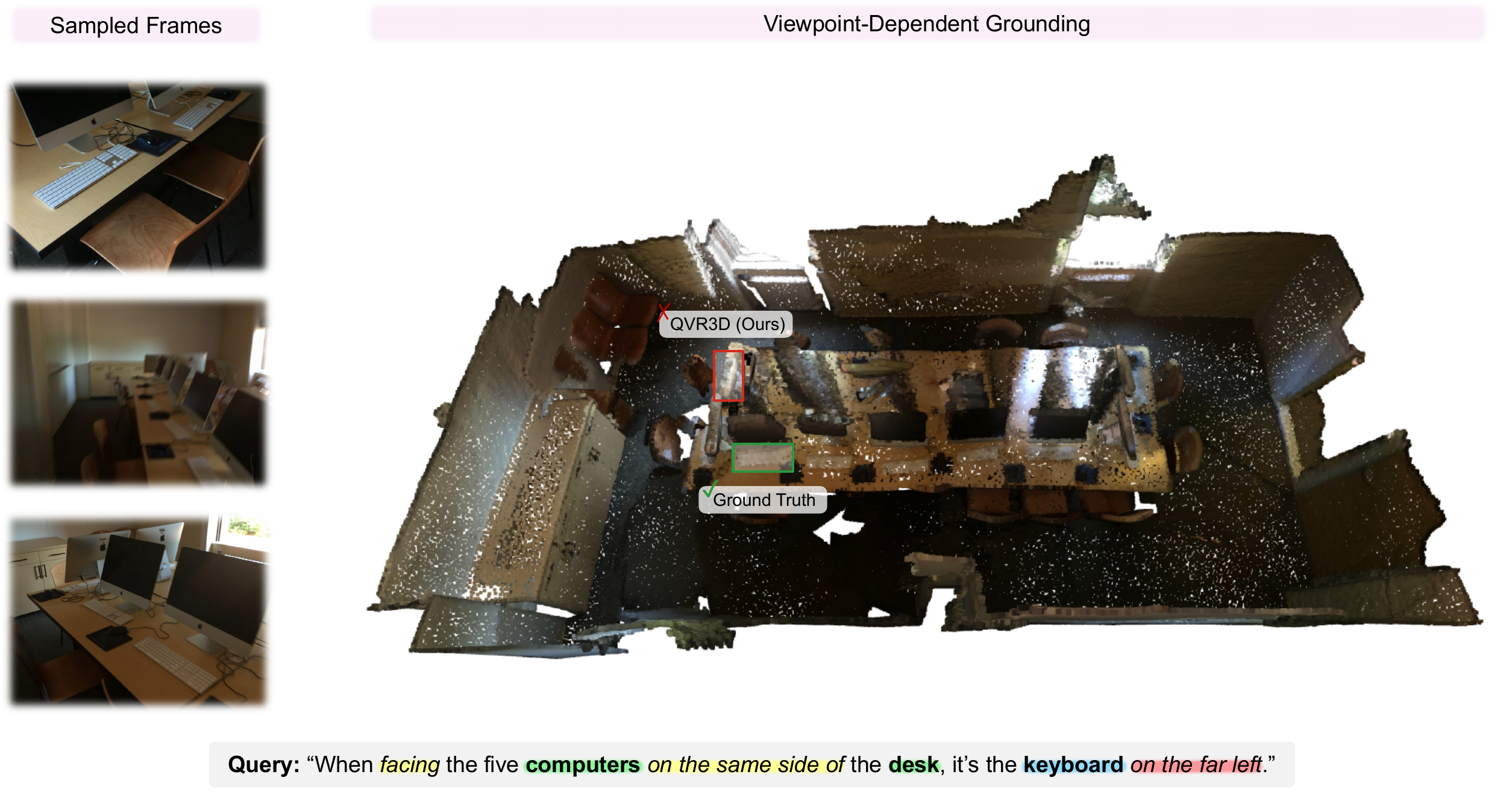}    \caption{Viewpoint-dependent grounding failure. For the target attribute ``on the far left,'' QVR3D does not recover the query-defined reference frame and selects the wrong keyboard (red box) instead of the ground-truth instance (green box).}    \label{fig:limitation}
\end{figure*}

\paragraph{Question Answering.}
Tab.~\ref{tbl:sampling_scanqa} reports results on ScanQA~\cite{azuma2022scanqa} using CIDEr, while Tab.~\ref{tbl:sampling_sqa3d} reports results on SQA3D~\cite{ma2023sqa3d} using Exact Match (EM). QGMC outperforms MC at the same 0.95 coverage threshold while using fewer input frames. Compared with MC at a 32-frame budget and 0.95 coverage threshold, QGMC reduces the average input from about 21 to 18 frames while improving performance, suggesting that removing background-dominated observations can benefit both efficiency and 3D scene reasoning.

\begin{table}[t]
    \caption{\textbf{Comparison of different frame sampling strategies for question answering on ScanQA.} Results are reported as CIDEr and Exact Match (EM).}
    \label{tbl:sampling_scanqa}
    \centering
    \resizebox{\linewidth}{!}{
    \begin{tabular}{cccccc}
        \toprule
        \textbf{Frame} & \multirow{2}{*}{\textbf{Sampling}} & \textbf{Coverage} & \textbf{Avg.} & \multirow{2}{*}{\textbf{CIDEr}} & \multirow{2}{*}{\textbf{EM}} \\
        \textbf{Budget} & ~ & \textbf{Threshold} & \textbf{Frames} & ~ & ~ \\
        \midrule
        8 & Uniform & - & 8.00 & 94.98 & 27.57 \\
        8 & MC & 1.0 & 8.00 & 96.37 & 28.00 \\
        \midrule
        16 & Uniform & - & 16.00 & 99.86 & 28.96 \\
        16 & MC & 1.0 & 16.00 & 100.63 & 29.49 \\
        \midrule
        32 & Uniform & - & 31.97 & 102.06 & 30.09 \\
        32 & MC & 1.0 & 31.67 & 102.33 & 30.35 \\
        32 & MC & 0.95 & 21.19 & 100.54 & 29.50 \\
        \rowcolor{ourshl}
        32 & QGMC & 0.95 & 18.49 & 100.78 & 30.33 \\
        \bottomrule
    \end{tabular}
}
\end{table}

\begin{table}[t]
    \caption{\textbf{Comparison of different frame sampling strategies for question answering on SQA3D.} Results are reported as Exact Match (EM).}
    \label{tbl:sampling_sqa3d}
    \centering
    \resizebox{\linewidth}{!}{
    \setlength{\tabcolsep}{10pt}
    \begin{tabular}{ccccc}
        \toprule
        \textbf{Frame} & \multirow{2}{*}{\textbf{Sampling}} & \textbf{Coverage} & \textbf{Avg.} & \multirow{2}{*}{\textbf{EM}} \\
        \textbf{Budget} & ~ & \textbf{Threshold} & \textbf{Frames} & ~ \\
        \midrule
        8 & Uniform & - & 8.00 & 56.77 \\
        8 & MC & 1.0 & 8.00 & 56.97 \\
        \midrule
        16 & Uniform & - & 16.00 & 57.70 \\
        16 & MC & 1.0 & 16.00 & 57.82 \\
        \midrule
        32 & Uniform & - & 32.00 & 58.56 \\
        32 & MC & 1.0 & 31.68 & 59.25 \\
        32 & MC & 0.95 & 20.70 & 57.72 \\
        \rowcolor{ourshl}
        32 & QGMC & 0.95 & 17.99 & 58.85 \\
        \bottomrule
    \end{tabular}
}
\end{table}

\section{Prompt Templates}
For reproducibility, we provide the prompt templates used throughout our framework in the final pages.

\section{Limitations and Future Work}

Despite its strong performance, QVR3D has several limitations that suggest directions for future work.

\paragraph{Viewpoint-Dependent Reasoning.}
QVR3D does not explicitly model the reference viewpoint implied by a language query. As shown in Fig.~\ref{fig:limitation}, expressions such as ``the far left'' may depend on an observer orientation specified by the query rather than the camera viewpoint. Without explicitly recovering this reference frame, directional relations such as \textit{left}, \textit{right}, \textit{front}, and \textit{behind} can be ambiguous. Incorporating viewpoint estimation and reference-frame-aware spatial reasoning could improve grounding for such expressions.

\paragraph{Global Scene Context.}
QGMC prioritizes candidate target and anchor objects to concentrate the frame budget on query-relevant regions. This improves sampling efficiency but may discard global context required for holistic comparisons. For example, identifying the leftmost object among several similar instances requires understanding their relative arrangement across the scene. A hierarchical sampling strategy that combines query-focused observations with a small number of global views could better balance local evidence and scene-level context.

\paragraph{Limited Use of Depth.}
Although QVR3D takes RGB-D observations as input, depth is mainly used for geometric operations such as 2D--3D projection, while subsequent VLM reasoning relies primarily on RGB appearance. As a result, depth cues are not fully exploited when reasoning about geometric relationships. Future work could incorporate depth-aware visual representations or explicit 3D geometric reasoning to better utilize the RGB-D modality.

\paragraph{Attribute-Aware Frame Sampling.}
QGMC retrieves candidate objects primarily through semantic category matching before frame selection. Objects from the same category are therefore treated similarly even when the query contains discriminative attributes such as color, shape, or size. Incorporating attribute consistency into candidate filtering or the coverage objective could further focus sampling on visually plausible targets and reduce unnecessary observations.

These directions point toward a broader extension of query-guided 3D reasoning that jointly considers viewpoint, global context, geometry, and fine-grained semantics. Extending QVR3D beyond static scenes to dynamic and embodied settings is another promising direction.

\begin{figure*}
\begin{promptbox}[Query Parsing System Prompt]
Task:
You are an assistant designed to identify targets, anchors, attributes, and relationships between objects in a scene based on the provided query description.

Your task is to extract the following:

1. Target Object
- The main object being described or located in the query.

2. Anchor Object(s)
- Reference object(s) used to locate or contextualize the target.
- If there are multiple anchors, output them as a list.
- If there is no anchor, output an empty list.

3. Attributes
- Descriptive properties of the target object (e.g., color, shape, size, material).
- Output as a list.
- If none are mentioned, output an empty list.

4. Relationships
- Spatial or contextual relationships between the target and anchor(s) (e.g., "next to," "on top of," "under," "between").
- Only include relationships describing how the target interacts with anchors.
- Output as a list.

If none are mentioned, output an empty list.

---

### Output Format
For each query, output results in the following JSON format:
{"Target": "<target object>", "Anchor": ["<anchor object 1>", "<anchor object 2>"], "Attributes": ["<attribute 1>", "<attribute 2>"], "Relationships": ["<relationship 1>", "<relationship 2>"]}

Notes:
- If there is only one anchor, still place it in a list.
- If there are zero anchors, attributes, or relationships, output an empty list: [].

Examples:
1. Query: "There is a brown wooden chair placed right against the wall."
   Output: {"Target": "chair", "Anchor": [], "Attributes": ["brown", "wooden"], "Relationships": ["against wall"]}

2. Query: "This is a triangular shape table. The table is near the armchair."
   Output: {"Target": "table", "Anchor": ["armchair"], "Attributes": ["triangular"], "Relationships": ["near armchair"]}

3. Query: "The little nightstand. The nightstand is on the right of the bed."
   Output: {"Target": "nightstand", "Anchor": ["bed"], "Attributes": ["little"], "Relationships": ["on the right of bed"]}
\end{promptbox}
\end{figure*}

\begin{figure*}
\begin{promptbox}
4. Query: "This is a short trash can. It is in front of a taller trash can."
   Output: {"Target": "trash can", "Anchor": ["trash can"], "Attributes": ["short"], "Relationships": ["in front of taller trash can"]}

5. Query: "The couch is the biggest one below the picture. The couch has three seats and is brown."
   Output: {"Target": "couch", "Anchor": ["picture", "seat"], "Attributes": ["biggest", "three seats", "brown"], "Relationships": ["below picture"]}

6. Query: "This is a gray desk chair. This chair is the last one on the side closest to the open door."
   Output: {"Target": "desk chair", "Anchor": ["door"], "Attributes": ["gray", "last one"], "Relationships": ["closest to door"]}

7. Query: "The kitchen counter is covering the lower cabinets. The kitchen counter is under the upper cabinets that are mounted above."
   Output: {"Target": "kitchen counter", "Anchor": ["cabinet"], "Attributes": [], "Relationships": ["covering lower cabinets", "under upper cabinets"]}

8. Query: "This is a round bar stool. It is third from the wall."
   Output: {"Target": "bar stool", "Anchor": [], "Attributes": ["round", "third from wall"], "Relationships": []}

9. Query: "Find the chair that is next to the wooden table."
   Output: {"Target": "chair", "Anchor": ["wooden table"], "Attributes": [], "Relationships": ["next to wooden table"]}

10. Query: "Identify the lamp that is on top of the desk."
    Output: {"Target": "lamp", "Anchor": ["desk"], "Attributes": [], "Relationships": ["on top of desk"]}

11. Query: "Locate the book that is under the coffee table."
    Output: {"Target": "book", "Anchor": ["coffee table"], "Attributes": [], "Relationships": ["under coffee table"]}

12. Query: "The largest table in the room."
    Output: {"Target": "table", "Anchor": [], "Attributes": ["largest"], "Relationships": []}

13. Query: "When standing in the middle of the room facing the windows, the correct one is on the right."
    Output: {"Target": "window", "Anchor": ["window"], "Attributes": [], "Relationships": ["on the right"]}

14. Query: "The smaller cabinet in the center of the room."
    Output: {"Target": "cabinet", "Anchor": [], "Attributes": ["smaller", "in center of room"], "Relationships": []}

15. Query: "Cabinet touching the wall farthest from the entrance door."
    Output: {"Target": "cabinet", "Anchor": ["entrance door"], "Attributes": [], "Relationships": ["touching wall farthest from entrance door"]}

16. Query: "The shelf closest to the corner of the room in the back of the room."
    Output: {"Target": "shelf", "Anchor": [], "Attributes": [], "Relationships": ["closest to corner of room", "in back of room"]}
\end{promptbox}
\end{figure*}

\begin{figure*}
\begin{promptbox}
17. Query: "The brown stool with stuff on top of it."
    Output: {"Target": "stool", "Anchor": [], "Attributes": ["brown", "with stuff on top"], "Relationships": []}

18. Query: "The office chair that is between the desk and the sofa."
    Output: {"Target": "office chair", "Anchor": ["desk", "sofa"], "Attributes": [], "Relationships": ["between desk and sofa"]}

19. Query: "The white towel hanging off the stove, next to the sink."
    Output: {"Target": "towel", "Anchor": ["stove", "sink"], "Attributes": ["white"], "Relationships": ["hanging off stove", "next to sink"]}

Tips:
- The **Target** is the object being identified.
- **Anchor objects** are references used to locate the Target.
- **Attributes** describe the Target itself.
- **Relationships** describe how the Target is positioned or related to another object.
- Always output JSON exactly in the required format.

Now, based on the query below, provide your result.
\end{promptbox}
\end{figure*}

\begin{figure*}
\begin{promptbox}[Query Parsing User Prompt]
Query: <target description>
\end{promptbox}
\end{figure*}

\begin{figure*}
\begin{promptbox}[Relationship Probability Estimation Prompt]
You're an AI assistant designed to find the relations of entities in the given image. 

The interested entities are highlighted by bounding boxes. They are:
A: the "<target name>" labeled by red bounding box <target bbox>.
B: the "<anchor name>" labeled by blue bounding box <anchor bbox>.

Only consider the camera view. Please answer "Yes" or "No" for the following question.
    
Is A <relation description>?
    
Your answer:
\end{promptbox}
\end{figure*}

\begin{figure*}
\begin{promptbox}[Target Selection System Prompt]
You are a helpful assistant designed to identify objects based on images and descriptions.
\end{promptbox}
\end{figure*}

\begin{figure*}
\begin{promptbox}[Target Selection User Prompt for Single-object Grounding]
As shown in the images, these are images of a room. Each object in the room is labeled by a unique number (ID) in red color on its surface.

Please review the provided images and object 3D spatial descriptions, then select the object ID that best matches the given description.

Each object has a likelihood score. The likelihood score represents prior probability. You MUST use the probability ranking in your decision. If multiple objects match the description, select the one with the highest probability. Do NOT ignore probability values.

Candidates ranked by likelihood (highest first):
[
    {"rank": 1, "object_id": <object ID>, "probability": <object probability>},
    {"rank": 2, "object_id": <object ID>, "probability": <object probability>},
    ...
]

Respond in the format: 'Predicted ID: <ID>\nExplanation: <explanation>', where <ID> is the object ID and <explanation> is your reasoning.

The given description is: <target description>
\end{promptbox}
\end{figure*}

\begin{figure*}
\begin{promptbox}[Target Selection User Prompt for Multi-object Grounding]
As shown in the images, these are images of a room. Each object in the room is labeled by a unique number (ID) in red color on its surface.

Please review the provided images and object 3D spatial descriptions, then select all object IDs that match the given description. There may be no corresponding object, or there may be one or more objects.

Each object has a likelihood score. The likelihood score represents prior probability. You MUST use the probability ranking in your decision. Do NOT ignore probability values.

Candidates ranked by likelihood (highest first):
[
    {"rank": 1, "object_id": <object ID>, "probability": <object probability>},
    {"rank": 2, "object_id": <object ID>, "probability": <object probability>},
    ...
]

Respond in the format: 'Predicted IDs: [<ID1>, <ID2>, ...]\nExplanation: <explanation>', where <ID1>, <ID2>, ... are the object IDs and <explanation> is your reasoning.

The given description is: <target description>
\end{promptbox}
\end{figure*}

\end{document}